\documentclass[]{spie}
\usepackage{graphicx,verbatim}
\usepackage{subcaption}
\usepackage{multirow}
\usepackage{booktabs}
\usepackage{array} 
\usepackage[xcdraw]{xcolor} 
\usepackage{colortbl} 
\usepackage[ruled,vlined]{algorithm2e}
\newcommand\rowSize{0.85cm}
\newcommand\rowColorOne{gray!10}
\newcommand\rowColorTwo{blue!10}
\newcommand\rowColorThree{yellow!10}
\usepackage{amsfonts}
\usepackage{physics}

\newcolumntype{P}[1]{>{\centering\arraybackslash}p{#1}}

\title{Monocular Marker-free Patient-to-Image Intraoperative Registration for Cochlear Implant Surgery}

\author[a]{Yike Zhang}
\author[a]{Eduardo Davalos Anaya}
\author[a,b]{Jack H. Noble}
\affil[a]{Dept. of Computer Science, Vanderbilt University}
\affil[b]{Dept. of Electrical and Computer Engineering, Vanderbilt University}

\begin{document}
\maketitle              
\begin{abstract}
This paper presents a novel method for monocular patient-to-image intraoperative registration, specifically designed to operate without any external hardware tracking equipment or fiducial point markers. Leveraging a synthetic microscopy surgical scene dataset with a wide range of transformations, our approach directly maps preoperative CT scans to 2D intraoperative surgical frames through a lightweight neural network for real-time cochlear implant surgery guidance via a zero-shot learning approach. Unlike traditional methods, our framework seamlessly integrates with monocular surgical microscopes, making it highly practical for clinical use without additional hardware dependencies and requirements. Our method estimates camera poses, which include a rotation matrix and a translation vector, by learning from the synthetic dataset, enabling accurate and efficient intraoperative registration. The proposed framework was evaluated on nine clinical cases using a patient-specific and cross-patient validation strategy. Our results suggest that our approach achieves clinically relevant accuracy in predicting 6D camera poses for registering 3D preoperative CT scans to 2D surgical scenes with an angular error within 10 degrees in most cases, while also addressing limitations of traditional methods, such as reliance on external tracking systems or fiducial markers.
\end{abstract}

\section{Introduction}
\begin{figure}[ht]
  \centering
  \resizebox{0.9\textwidth}{!}{%
  \begin{subfigure}{0.23\textwidth}
        \centering
        \includegraphics[width=0.9\textwidth]{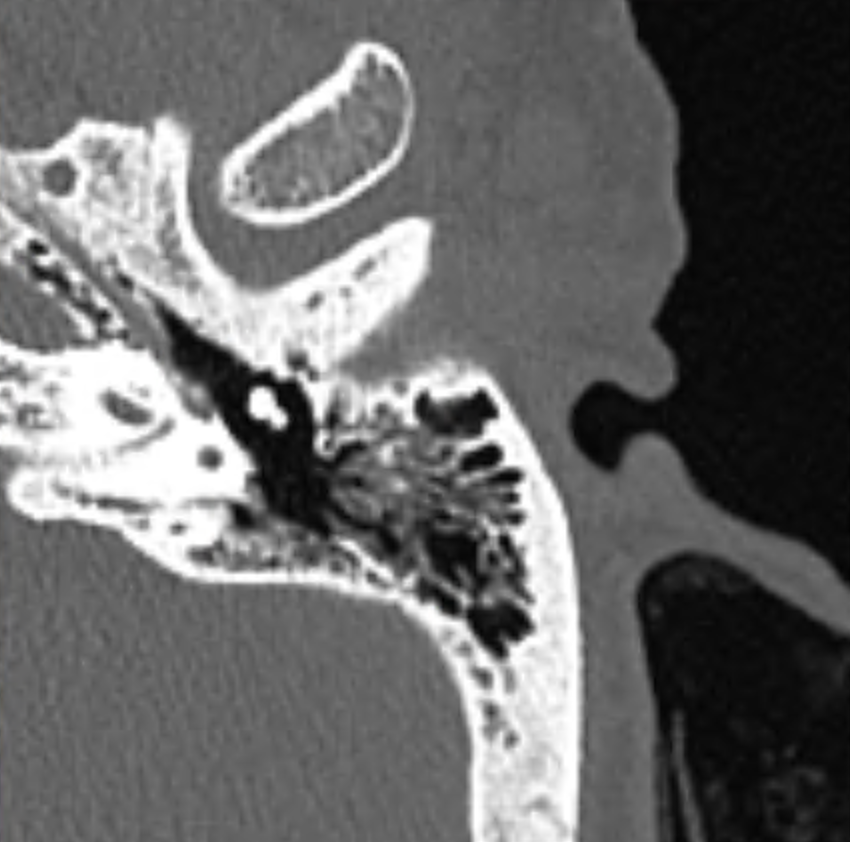}
        \caption{\footnotesize{Original CT}}
        \label{fig:preop}
    \end{subfigure}
    \hfill
    \begin{subfigure}{0.23\textwidth}
        \centering
        \includegraphics[width=0.9\textwidth]{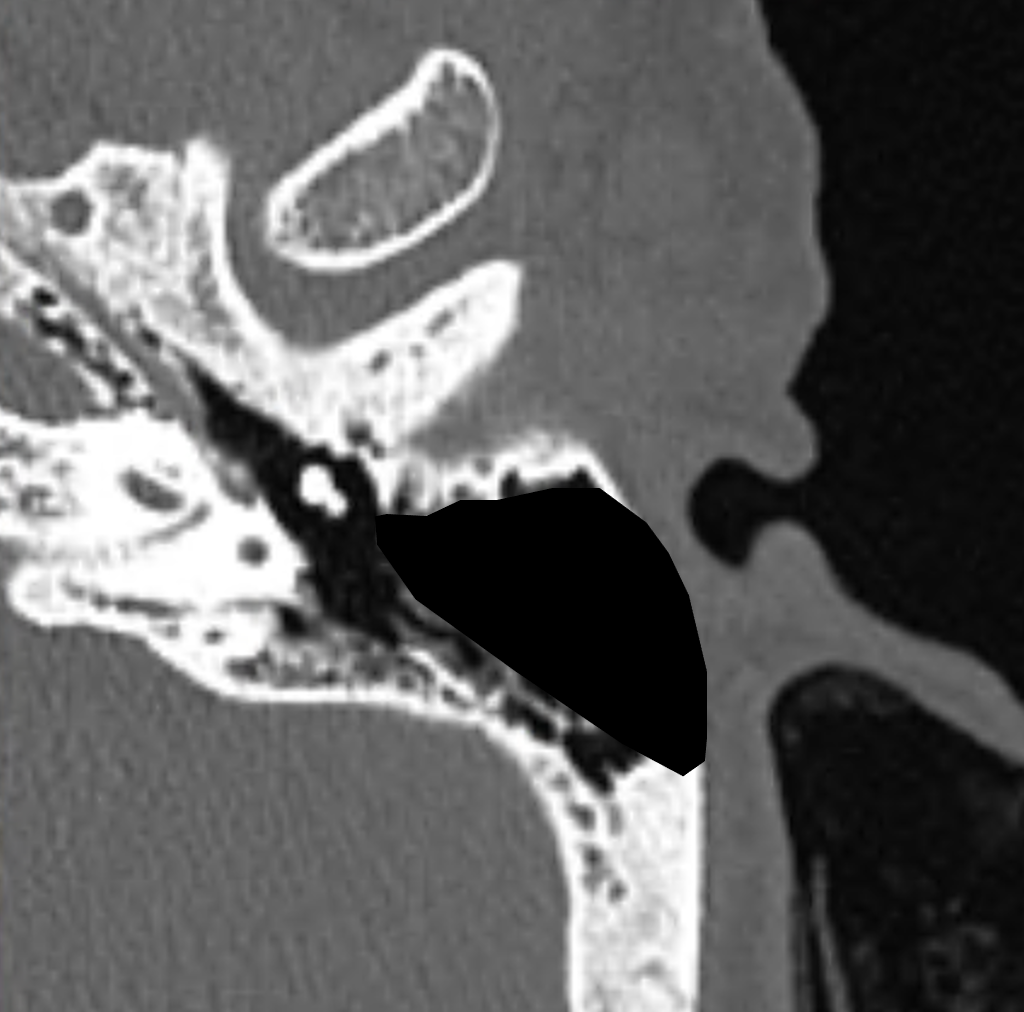}
        \caption{\footnotesize{Synthetic CT}}
        \label{fig:synthesized_preop}
    \end{subfigure}
    \hfill
    \begin{subfigure}{0.23\textwidth}
        \centering
        \includegraphics[width=0.9\textwidth]{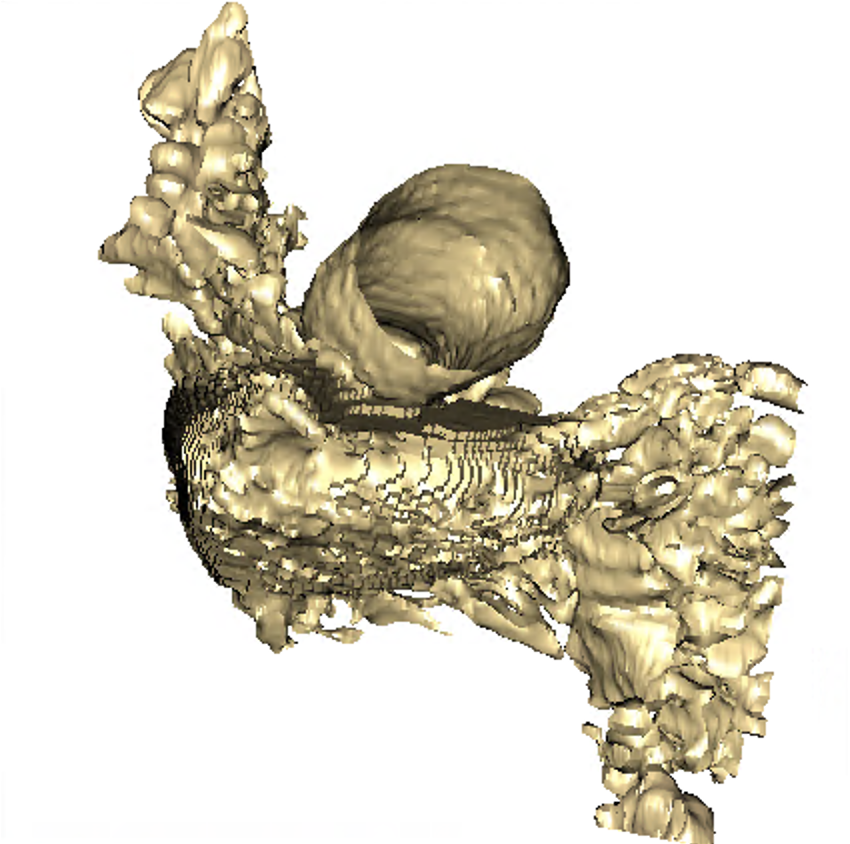}
        \caption{\footnotesize{Constructed CT}}
        \label{fig:reconstructed_surface}
    \end{subfigure}
    \hfill
    \begin{subfigure}{0.23\textwidth}
        \centering
        \includegraphics[width=0.9\textwidth]{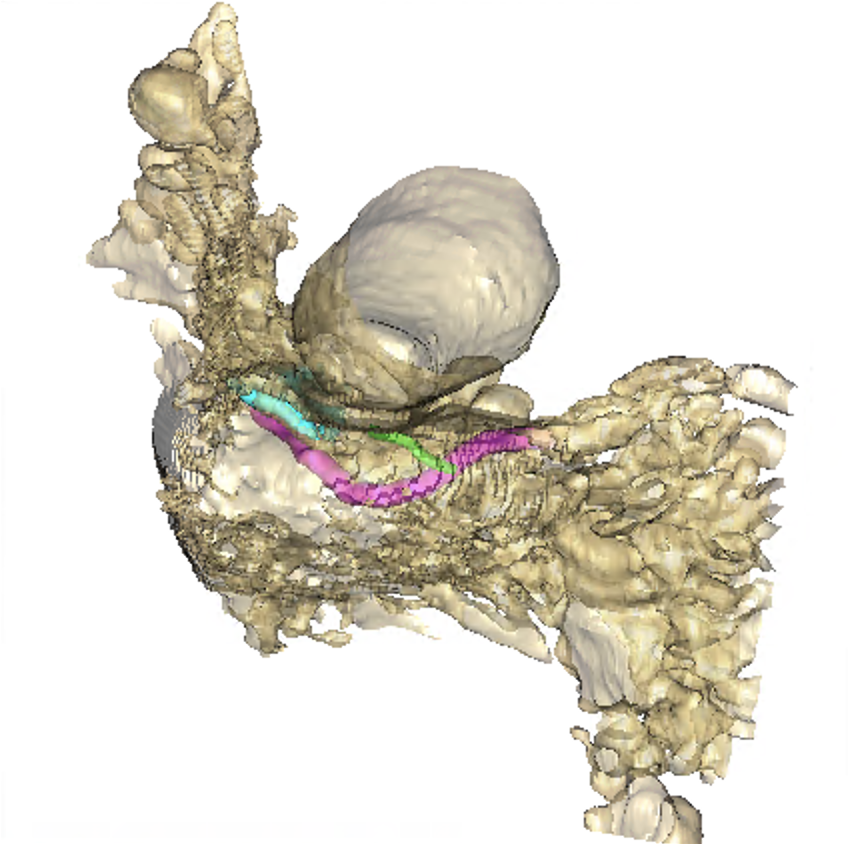}
        \caption{\footnotesize{Overlay}}
        \label{fig:ear_structures_overlay}
    \end{subfigure}%
    }
\caption{\textbf{Construct Postmastoidectomy Surface}. Extract the 3D surface directly from the preoperative CT scan.}
\label{fig:recontruct_surface}
\end{figure}

Cochlear Implant (CI) surgeries are specialized procedures designed to restore hearing in individuals with moderate-to-profound hearing loss, significantly improving their quality of life \cite{labadie2018preliminary}. This paper addresses a critical challenge in image-guided cochlear implant surgery: directly mapping 3D anatomical surfaces extracted from preoperative CT scans to intraoperative 2D surgical microscopy scenes for accurate registration. By leveraging popular deep-learning-based models, we predict the camera poses of the entire CT surface using fully synthesized surgical scenes and apply the trained model to real surgical data via zero-shot approach. To achieve this, we integrate methodologies from previous research (e.g., \cite{zhang2024mmunsupervisedmambabasedmastoidectomy}, \cite{zhang2024mastoidectomymultiviewsynthesissingle}, and \cite{zhang2025ssddgansinglestepdenoisingdiffusion}) to build a first-of-its-kind synthetic CI surgery dataset, and propose a lightweight framework for real-time intraoperative patient-to-image registration. In CI procedures, preoperative CT scans are registered with intraoperative microscopy views of the post-mastoidectomy surface to guide otologists in determining the optimal electrode array insertion angle \cite{10.1117/12.3008830}.
Accurate electrode placement within the critical 10-degree angular distance threshold is essential for minimizing the risk of basilar membrane damage, enhancing surgical safety, and reducing postoperative complications \cite{labadie2018preliminary}. The accuracy of electrode positioning is correlated with a patient’s likelihood of achieving optimal hearing restoration, making the 3D-to-2D intraoperative registration a vital component of successful cochlear implant surgery.
Intraoperative patient-to-image registration is a long-standing challenge in image-guided surgery \cite{tawfik2024cochlearimplantationslimprecurved}, requiring precise alignment of the patient’s anatomy with intraoperative imaging modalities such as MRI, CT, or ultrasound. Image-guided surgery can provide information that assists navigation and tracking, potentially improving surgical outcomes, and being utilized as a valuable training tool for junior surgeons. Traditional registration methods often rely on fiducial markers, such as screws or beads, attached to the patient’s anatomy \cite{lin2023modern}. These markers enable accurate alignment among different modalities. However, they are inherently invasive, time-consuming, and require additional setup that affects surgical workflow efficiency, posing challenges for resource-limited hospitals.
Recently, \cite{Furuse2023-un} presented a study about a marker-based approach for brain tumor surgeries. Their study evaluated the impact of surgical positioning-induced skin distortion on the clinical accuracy of a navigation system. Another example of the marker-based approach is the use of NDI (Northern Digital Inc.) optical navigation tracking systems, which are widely adopted in neurosurgery and orthopedics \cite{10.3389/fnbot.2021.636772,healthcare10101815,sahovaler2022automatic}. 
While marker-based registration techniques have been widely used in image-guided surgeries, their reliance on fiducial markers and external tracking systems makes them disruptive and resource-intensive \cite{taleb2023image}, limiting their applicability to CI surgery. Recent developments in marker-free registration have demonstrated the potential of using deep learning-based methods in aligning 3D anatomical structures to 2D surgical scenes.
For example, \cite{10.1117/12.3008830} attempted to directly register anatomical structures, such as the ossicles, to 2D surgical scenes in cochlear implant surgery. However, the high occlusion rate of the ossicles posed significant challenges, resulting in suboptimal pose predictions with an average angular distance error of approximately 20\textdegree, which is inadequate for reliable applications in this surgical domain. Another approach to the patient-to-image intraoperative registration, proposed by \cite{haouchine2023learningexpectedappearancesintraoperative}, focuses specifically on applications in neurosurgery. Their method leverages preoperative imaging to synthesize patient-specific expected views across a range of transformations, effectively avoiding issues caused by noisy intraoperative images.
Directly aligning preoperative data to the monocular surgical microscope views without any external equipment in the context of CI surgery is challenging. Current patient-to-image intraoperative registration approaches either focus on different surgical domains \cite{haouchine2023learningexpectedappearancesintraoperative,Haouchine2022} or lack the sufficient datasets required to train such neural networks. This gap highlights the need for approaches that can leverage preoperative data to accurately synthesize intraoperative views, enabling direct patient-to-image alignment without external equipment.
\cite{zhang2024mmunsupervisedmambabasedmastoidectomy} proposed a technique to predict the mastoidectomy shape from preoperative CT scans, allowing for the construction of the postmastoidectomy surface visible in the surgical microscope via iso-surfacing, as illustrated in Figure~\ref{fig:preop} \ref{fig:synthesized_preop}, and \ref{fig:reconstructed_surface}.
Key ear structures, including the ossicles (cyan), facial nerve (magenta), and chorda tympani (lime), are overlaid on the constructed postmastoidectomy surface in Figure~\ref{fig:ear_structures_overlay}.
Since their relative positions remain fixed in preoperative CT scans, accurately registering the postmastoidectomy surface intraoperatively will inherently align all associated anatomical structures from the CT scan, including those hidden beneath the visible surgical surface. This method also paves the way for multi-view surgical scene synthesis, as proposed in \cite{zhang2024mastoidectomymultiviewsynthesissingle}, generating synthetic intraoperative postmastoidectomy surface scenes to assist CI surgical navigation and analysis.
Our \textbf{contributions} are listed below:
\begin{itemize}
    \item We are the first to implement real-time, marker-free intraoperative registration for CI surgery, leveraging a lightweight deep learning-based approach to accurately align 3D patient anatomical structures with 2D monocular surgical microscope scenes at approximately 40 frames per second on an NVIDIA GeForce RTX 4090 GPU.
    \item We build a comprehensive synthetic surgical scene dataset with known camera poses for each frame. To the best of our knowledge, this is the first dataset of its kind specifically designed for CI surgery.
    \item We evaluate our approach using both patient-specific validation, ensuring effectiveness for individual patients, and cross-patient validation, demonstrating robustness and generalizability to unseen patients.
\end{itemize}
\section{Method}
\subsection{Problem Formulation}
In our proposed method, we have the postmastoidectomy surfaces $M = (V, F)$ extracted from the preoperative CT scans, the synthetic 2D monocular surgical microscope images $\mathbf{I} \in \mathbb{R}^{H \times W}$ and their corresponding camera extrinsic matrix $\mathbf{P} = [\mathbf{R}|\vec{t}]$. We define a set of 3D vertices $V = \{\vec{v}_i \in \mathbb{R}^{N\times3}\}$ and a set of triangle faces $F = \{\vec{f}_i \in \mathbb{Z}^{N'\times3}\}$, where $N$ and $N'$ represents the number of vertices and faces, respectively. $H$ and $W$ denote the height and the width of $\mathbf{I}$, $\mathbf{R} \in SO(3)$ represents the rotation matrix, and $\vec{t} \in \mathbb{R}^3$ is the translation vector. We aim to use a neural network parameterized by $\theta$ to directly regress the camera pose $\mathbf{P}' = [\mathbf{R}'|\vec{t}']$, as the process can be denoted as $f_\theta(\mathbf{I})$. 
Generally, in a camera pose $\mathbf{P}$, $\mathbf{R}$ can be represented with quaternions, Axis angles, or Euler angles. However, as suggested by \cite{zhou2020continuityrotationrepresentationsneural}, all representations for the rotation matrix of four or fewer dimensions in the real Euclidean spaces are discontinuous and difficult for the gradients to back-propagate. Moreover, the discontinuity often leads to a significantly large error when regressing rotation parameters. Therefore, we adopt the continuous 6D rotation representations for $\mathbf{R}$ and $\mathbf{R'}$. Suppose $\mathbf{R} \in \mathbb{R}^{3 \times 3}$ and $\mathbf{R} = \left[ \begin{array}{c|c|c} \vec{r}_1 & \vec{r}_2 & \vec{r}_3 \end{array} \right]$, the 6D representation $\mathbf{\hat{R}}$ is defined as the combination of the first two columns of $\mathbf{R}$: $\mathbf{\hat{R}} = \left[ \begin{array}{c|c} \vec{r}_1 & \vec{r}_2\end{array} \right]$, where the $2 \times 3$ matrix is flattened to a $6 \times 1$ vector. The rotation matrix $\mathbf{R}$ can be recovered by computing the following equations given the $\vec{r}_1$ and $\vec{r}_2$ of $\mathbf{\hat{R}}$:
\begin{align}
\left\{ 
\begin{aligned}
& \vec{r}_1'= \mu\left(\vec{r}_1\right) \\
& \vec{r}_2' = \mu\left(\vec{r}_2 - (\vec{r}_1' \boldsymbol{\cdot} \vec{r}_2\right)\vec{r}_1' ) \\
& \vec{r}_3' = \vec{r}_1' \times \vec{r}_2'
\end{aligned} 
\right.
\end{align} where $\mu(\cdot)$ represents the vector normalization. Finally, $\mathbf{R}$ can be reconstructed as $\mathbf{R} = \left[ \begin{array}{c|c|c} \vec{r}_1' & \vec{r}_2' & \vec{r}_3' \end{array} \right]^T$. The rotation error, $d_{rot}$ between $\mathbf{R}$ and $\mathbf{R}'$ is defined in Equation~\ref{eq: d_rot}, and we also compute $d_{rot'} = \norm{\mathbf{R}' - \mathbf{R}}_2^2$. Similarly, the translational error, $d_{t}$ between $\vec{t}$ and $\vec{t}'$ is given by $d_{t} = \norm{\vec{t}' - \vec{t}}_2^2$.
\begin{equation}
    d_{rot} = \arccos\left(\frac{\mathrm{Tr}(\mathbf{R}^T\mathbf{R}') - 1}{2}\right)
\label{eq: d_rot}
\end{equation}
To integrate these errors during neural network backpropagation, we define a weighted loss function $\mathcal{L}_\theta$, as described in Equation~\ref{eq: combined_loss}:
\begin{equation}
    \mathcal{L}_\theta = \lambda_{rot} \cdot d_{rot}(\mathbf{R}', \mathbf{R}) + \lambda_{rot'} \cdot d_{rot'}(\mathbf{R}', \mathbf{R}) + \lambda_{t} \cdot d_{t}(\vec{t}', \vec{t}),
\label{eq: combined_loss}
\end{equation} where $\lambda_{rot}$, $\lambda_{rot'}$, and $\lambda_{t}$ are weighting factors for rotation and translation errors, respectively. The objective function $\mathcal{O}$ in Equation~\ref{eq:objective_function} is to minimize the loss over the synthetic training dataset:
\begin{equation}
    \mathcal{O} = \min_{\theta}\frac{1}{N}\sum_{i=1}^{N}\mathbf{L}(\theta; \mathbf{I}_i, \mathbf{P}_i),
\label{eq:objective_function}
\end{equation} where $N$ is the number of training samples, $\mathbf{I}_i$ and $\mathbf{P}_i$ represent the input 2D image and corresponding ground truth pose for the $i$-th training sample.
\subsection{Synthetic Dataset Generation}
\begin{figure}[ht]
  \centering
  \resizebox{0.9\textwidth}{!}{%
  \begin{subfigure}[b]{0.23\textwidth}
      \centering
      \includegraphics[width=\textwidth]{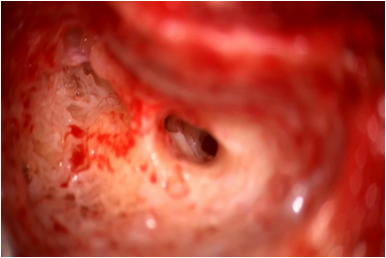}
      \caption{Surgical Scene}
      \label{fig:frame_168}
  \end{subfigure}
  \hfill
  \begin{subfigure}[b]{0.23\textwidth}
      \centering
      \includegraphics[width=\textwidth]{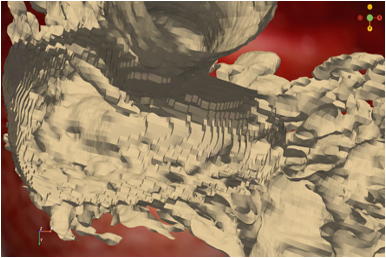}
      \caption{Registration}
      \label{fig:frame_168_2d_3d}
  \end{subfigure}
  \hfill
  \begin{subfigure}[b]{0.23\textwidth}
      \centering
      \includegraphics[width=\textwidth]{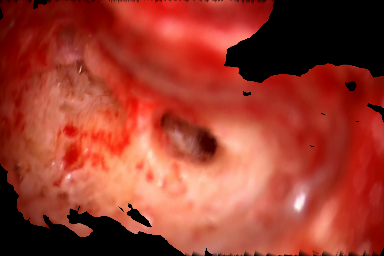}
      \caption{Texturing}
      \label{fig:frame_168_pyvista_masked}
  \end{subfigure}
  \hfill
  \begin{subfigure}[b]{0.23\textwidth}
      \centering
      \includegraphics[width=\textwidth]{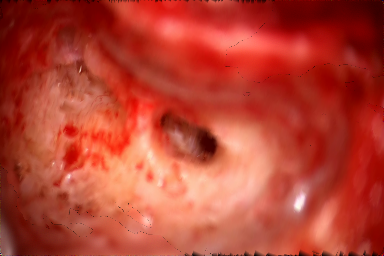}
      \caption{Synthetic Scene}
      \label{fig:frame_168_pyvista_reconstructed}
  \end{subfigure}%
  }
  \caption{\textbf{Pipeline for Synthesizing a Postmastoidectomy Surgical Scene}. (a) original surgical scene, (b) 3D-to-2D registration, (c) textured postmastoidectomy surface, and (d) synthetic surgical scene.}
  \label{fig:pipeline}
\end{figure}
\begin{figure}[ht]
  \centering
  \begin{minipage}[b]{0.15\textwidth}
      \centering
      \includegraphics[width=\textwidth]{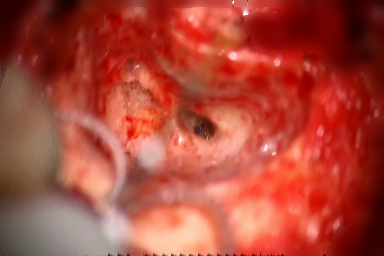}
  \end{minipage}
  \begin{minipage}[b]{0.15\textwidth}
      \centering
      \includegraphics[width=\textwidth]{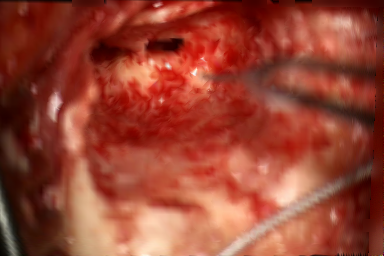}
  \end{minipage}
  \begin{minipage}[b]{0.15\textwidth}
      \centering
      \includegraphics[width=\textwidth]{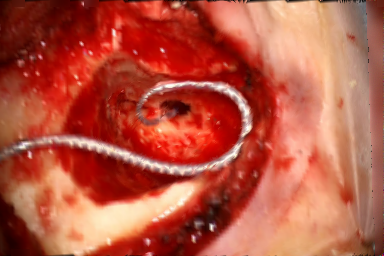}
  \end{minipage}
  \begin{minipage}[b]{0.15\textwidth}
      \centering
      \includegraphics[width=\textwidth]{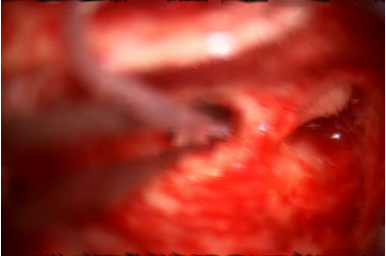}
  \end{minipage}
  \begin{minipage}[b]{0.15\textwidth}
      \centering
      \includegraphics[width=\textwidth]{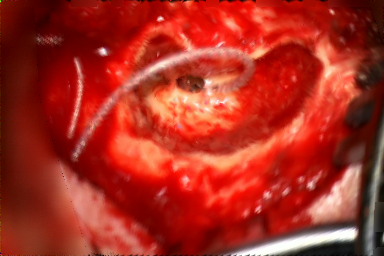}
  \end{minipage}
  \begin{minipage}[b]{0.15\textwidth}
      \centering
      \includegraphics[width=\textwidth]{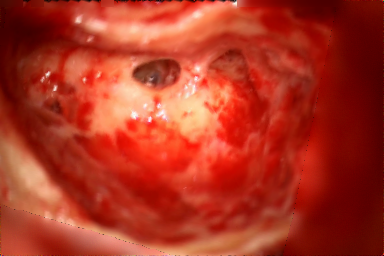}
  \end{minipage}
  \begin{minipage}[b]{0.15\textwidth}
      \centering
      \includegraphics[width=\textwidth]{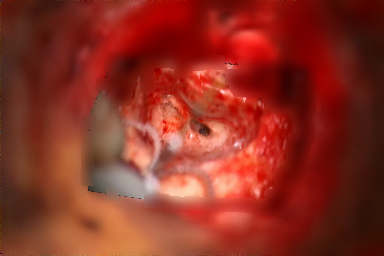}
  \end{minipage}
  \begin{minipage}[b]{0.15\textwidth}
      \centering
      \includegraphics[width=\textwidth]{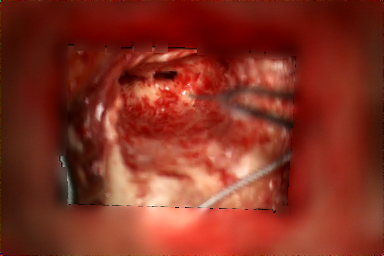}
  \end{minipage}
  \begin{minipage}[b]{0.15\textwidth}
      \centering
      \includegraphics[width=\textwidth]{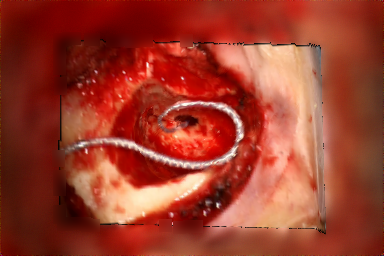}
  \end{minipage}
  \begin{minipage}[b]{0.15\textwidth}
      \centering
      \includegraphics[width=\textwidth]{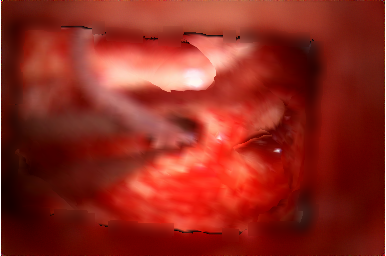}
  \end{minipage}
  \begin{minipage}[b]{0.15\textwidth}
      \centering
      \includegraphics[width=\textwidth]{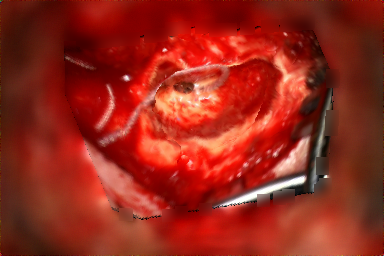}
  \end{minipage}
  \begin{minipage}[b]{0.15\textwidth}
      \centering
      \includegraphics[width=\textwidth]{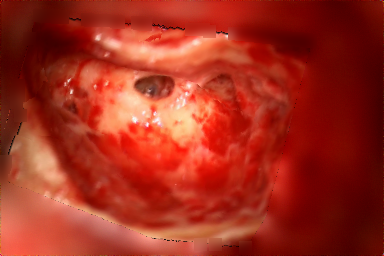}
  \end{minipage}
  \begin{minipage}[b]{0.15\textwidth}
      \centering
      \includegraphics[width=\textwidth]{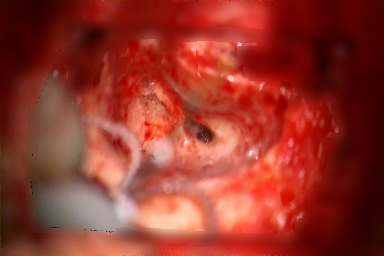}
      \footnotesize{Case 1}
  \end{minipage}
  \begin{minipage}[b]{0.15\textwidth}
      \centering
      \includegraphics[width=\textwidth]{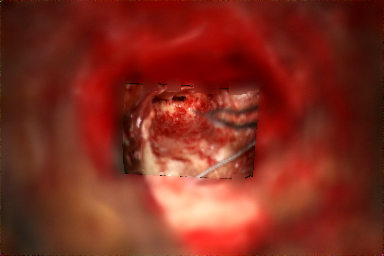}
      \footnotesize{Case 2}
  \end{minipage}
  \begin{minipage}[b]{0.15\textwidth}
      \centering
      \includegraphics[width=\textwidth]{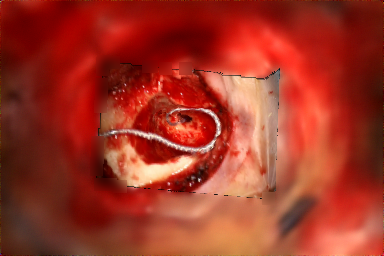}
      \footnotesize{Case 3}
  \end{minipage}
  \begin{minipage}[b]{0.15\textwidth}
      \centering
      \includegraphics[width=\textwidth]{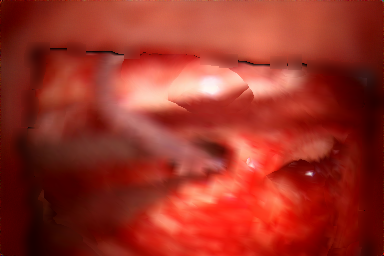}
      \footnotesize{Case 4}
  \end{minipage}
  \begin{minipage}[b]{0.15\textwidth}
      \centering
      \includegraphics[width=\textwidth]{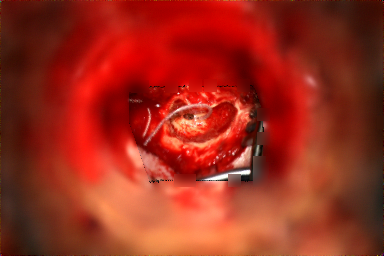}
      \footnotesize{Case 5}
  \end{minipage}
  \begin{minipage}[b]{0.15\textwidth}
      \centering
      \includegraphics[width=\textwidth]{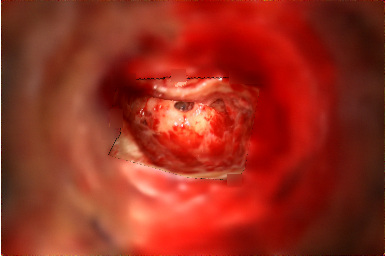}
      \footnotesize{Case 6}
  \end{minipage}
  
  \caption{\textbf{Multi-view Synthetic Surgical Scenes}. Each synthetic surgical scene was generated using distinct $\mathbf{P}$, to highlight the variability and effectiveness of our approach in capturing diverse perspectives during the multi-view surgical scene synthesis across both the left and right ears of different patients.}
  \label{fig:multi-view}
\end{figure}

An overview of the process for synthesizing the surgical surface is presented in Figure~\ref{fig:pipeline}. The postmastoidectomy surfaces of all patients, predicted from preoperative CT scans, are pre-registered to an atlas mesh in canonical space. An initial pose $\mathbf{P}_0$ of the atlas mesh is obtained using Vision6D in Figure~\ref{fig:frame_168_2d_3d}. The postmastoidectomy surface of the atlas mesh is then textured using $\mathbf{P}_0$ and the given camera intrinsics matrix $\mathbf{K}(f_x = 18466\text{mm}, f_y = 19172\text{mm})$, following the method proposed in \cite{zhang2024mastoidectomymultiviewsynthesissingle}, as demonstrated in Figure~\ref{fig:frame_168_pyvista_masked}. Subsequently, the missing regions of the surgical scene are completed using the SSDD-GAN method \cite{zhang2025ssddgansinglestepdenoisingdiffusion} in Figure~\ref{fig:frame_168_pyvista_reconstructed}. By applying a set of transformations $T = \{\mathbf{T}_0,..., \mathbf{T}_N\}$ to $\mathbf{P}_0$ while ensuring the rotation matrices are centered on the mesh centroid, we generate a comprehensive synthetic CI surgical scene dataset $(P = \{\mathbf{P}_0,...,\mathbf{P}_N\}, I=\{\mathbf{I}_0,...,\mathbf{I}_N\})$ that simulates various view angles and produces multi-view surgical scenes with VTK-based PyVista rendering library \cite{sullivan2019pyvista}. The new pose $\mathbf{P}_i$ for each synthetic view is given by $\mathbf{P}_i = \mathbf{T}_i \times \mathbf{P}_0$.
The multi-view synthetic surgical scenes with different ground truth $\mathbf{P}$ are illustrated in Figure~\ref{fig:multi-view}.
\subsection{Model Architecture}
\begin{figure}[ht]
    \centering
    \includegraphics[width=0.9\linewidth]{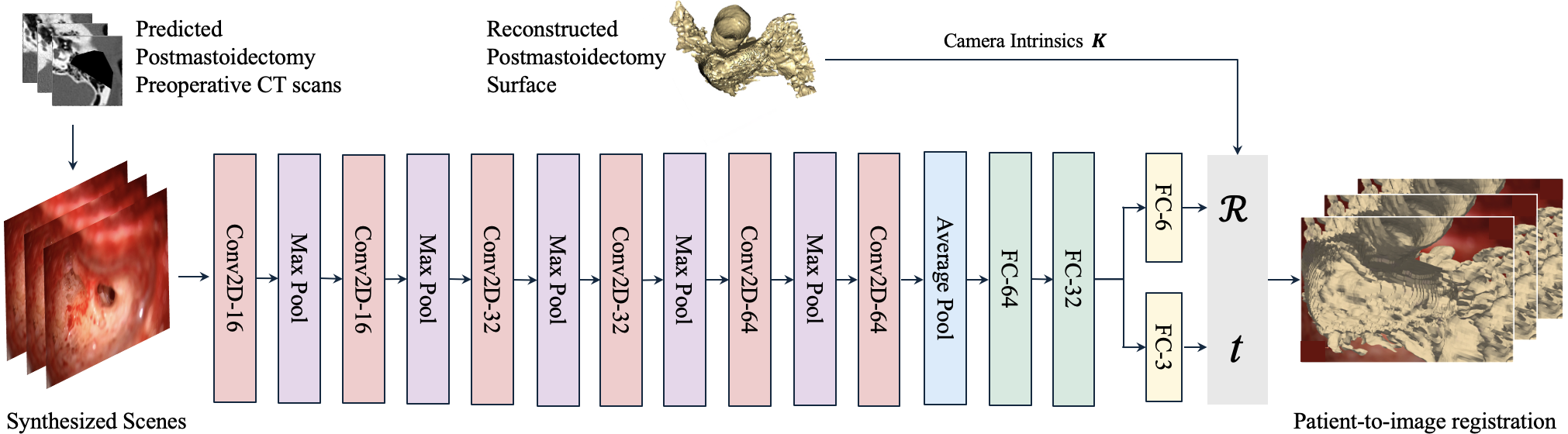}
    \caption{\textbf{Proposed Pose Regression Model}. $\vec{R}$ and $\vec{t}$ are output by the network to register the 3D postmastoidectomy surface directly to the 2D image.}
    \label{fig:framework}
\end{figure}

To optimize the objective function, we develop a lightweight CNN-based \cite{oshea2015introductionconvolutionalneuralnetworks} pose regression model $f_\theta$, specifically designed for the real-time 3D-to-2D intraoperative registration task. We add Dropout layers \cite{dropout} and LeakyReLU activation functions \cite{xu2015empiricalevaluationrectifiedactivations} inside the layers to improve model robustness and prevent overfitting. The model shown in Figure~\ref{fig:framework} is trained exclusively on a synthetic surgical dataset with pre-defined camera poses $\mathbf{P}$.
The proposed network is optimized to predict the camera poses $\mathbf{P}'$ for input images $\mathbf{I}$ by updating the network parameters $\theta$. To address potential challenges such as occlusions and varying lighting conditions that may occur in the surgical scene that are commonly caused by factors like surgeon hand movements, surgical tool dynamics, or sudden changes in the operating environment, we incorporate a range of data augmentation techniques. These include grid masking \cite{chen2024gridmaskdataaugmentation} to simulate occluded regions and brightness adjustments to account for different lighting, ensuring that the model is robust to diverse real-world scenarios that may encountered during surgery.
\section{Results}
\begin{figure}[ht]
  \centering
  \resizebox{0.9\textwidth}{!}{%
  \begin{minipage}{0.35\textwidth}
        \centering
        \begin{minipage}[c]{0.05\textwidth}
            \centering
            \rotatebox{90}{\footnotesize{patient-specific}}
        \end{minipage}
        \begin{minipage}[c]{0.9\textwidth}
        \includegraphics[width=0.7\textwidth]{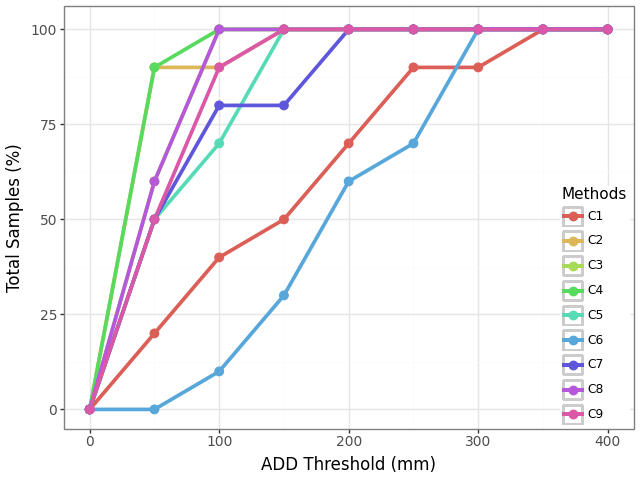}
        \end{minipage}

        \begin{minipage}[c]{0.05\textwidth}
            \centering
            \rotatebox{90}{\footnotesize{cross-patient}}
        \end{minipage}
        \begin{minipage}[c]{0.9\textwidth}
        \includegraphics[width=0.7\textwidth]{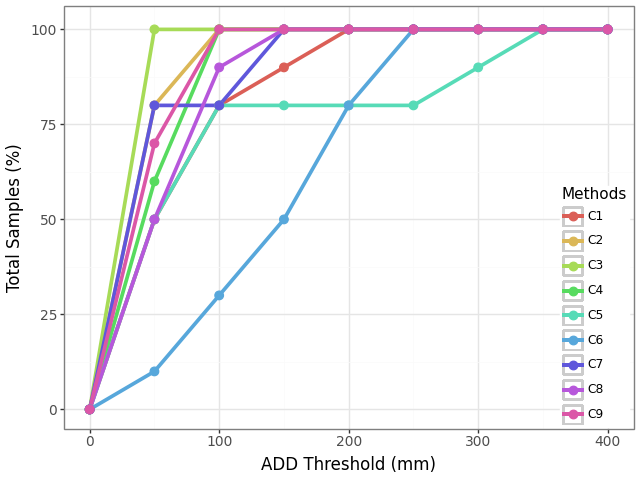}
        \end{minipage}

        \footnotesize{ADD Metric}
    \end{minipage}
    \begin{minipage}{0.35\textwidth}
        \centering
        \begin{minipage}[c]{0.05\textwidth}
            \centering
            \rotatebox{90}{\footnotesize{patient-specific}}
        \end{minipage}
        \begin{minipage}[c]{0.9\textwidth}
        \includegraphics[width=0.7\textwidth]{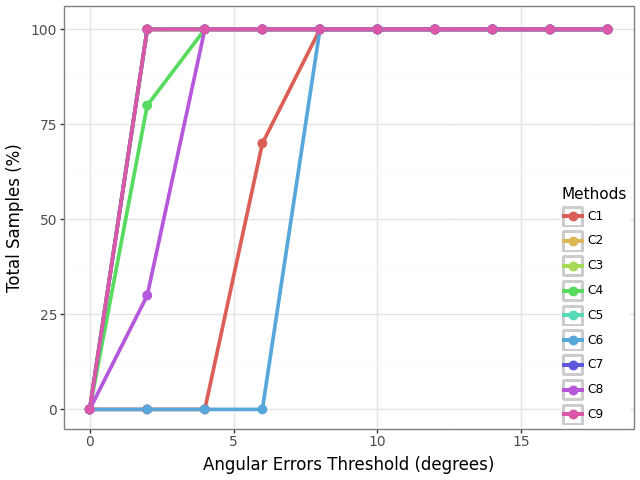}
        \end{minipage}
        
        \begin{minipage}[c]{0.05\textwidth}
            \centering
            \rotatebox{90}{\footnotesize{cross-patient}}
        \end{minipage}
        \begin{minipage}[c]{0.9\textwidth}
        \includegraphics[width=0.7\textwidth]{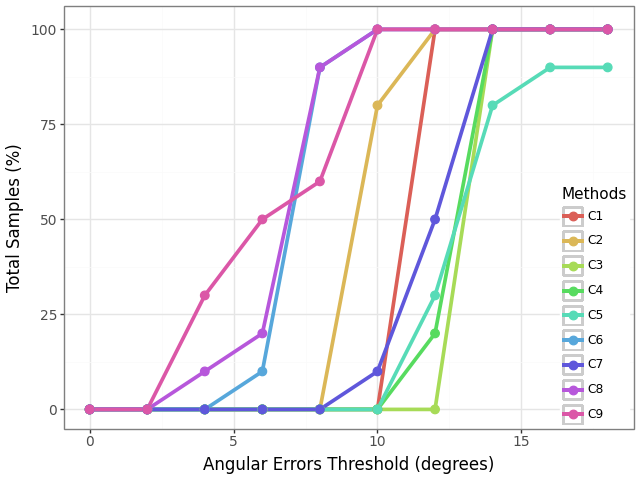}
        \end{minipage}

        \footnotesize{Rotation Metric}
    \end{minipage}
    \begin{minipage}{0.35\textwidth}
        \centering
        \begin{minipage}[c]{0.05\textwidth}
            \centering
            \rotatebox{90}{\footnotesize{patient-specific}}
        \end{minipage}
        \begin{minipage}[c]{0.9\textwidth}
        \includegraphics[width=0.7\textwidth]{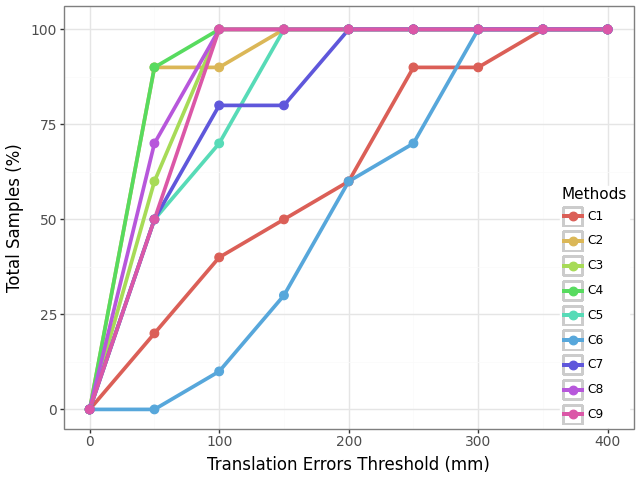}
        \end{minipage}
        
        \begin{minipage}[c]{0.05\textwidth}
            \centering
            \rotatebox{90}{\footnotesize{cross-patient}}
        \end{minipage}
        \begin{minipage}[c]{0.9\textwidth}
        \includegraphics[width=0.7\textwidth]{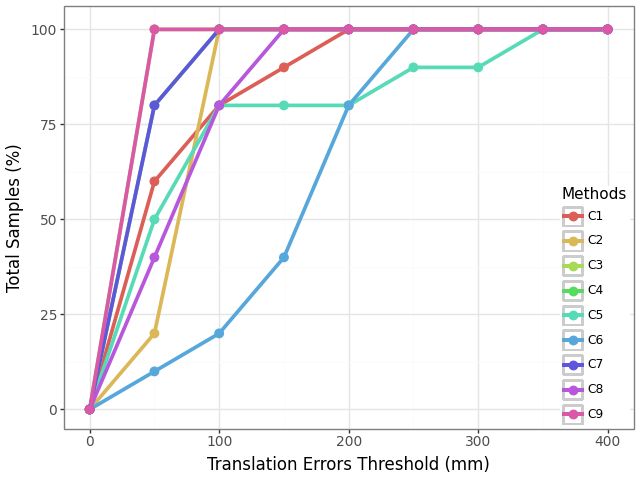}
        \end{minipage}

        \footnotesize{Translation Metric}
    \end{minipage}%
    }
\caption{\textbf{Performance Comparisons.} Comprehensive evaluation of nine individual patient cases under both patient-specific and cross-patient scenarios to analyze robustness and generalizability.}
\label{fig:performance_overall_comparison}
\end{figure}

We synthesized 10,000 frames with varying $\mathbf{P}$ for each of the nine clinical surgery videos, with each video corresponding to a distinct patient. For comparison, we manually annotated a total of 90 poses, with 10 poses per patient across 9 patients, each paired with its corresponding surgical frame. Table~\ref{tab:quantitative_evaluation} summarizes the model's performance on real datasets, evaluated using two validation strategies: a standard train-validation-test split for patient-specific assessment and a computationally intensive leave-one-out cross-validation to assess cross-patient generalizability. \textit{Patient-specific} refers to models trained on synthetic data ($\sim$10-minute training) and tested on real surgical data for the same patient. 
In contrast, \textit{Cross-patient} assesses generalizability by excluding one patient case during training and testing on their real images. For our quantitative evaluation, we use the Average Distance Metric \cite{6619221} (ADD) (in mm), which calculates the mean pairwise distance between the 3D surface vertices transformed by $\mathbf{P}$ and $\mathbf{P}'$. Rotation errors $E_{rot}$ are quantified using angular distances in degrees, while translation errors $E_{t}$ are obtained using the Euclidean distance metric in millimeters to capture positional discrepancies. For the \textit{patient-specific}, the average ADD, $E_{rot}$, and $E_{t}$ are 74.58 mm, 2.42$^\circ$, and 74.48 mm, respectively. For the \textit{cross-patient}, they are 61.31 mm, 10.21$^\circ$, and 52.99 mm across all cases. ADD and $E_{t}$ was predominantly contributed by the z-translation error in most cases, indicating that most of the translational offset occurred along the depth axis. 
The patient-specific results indicate that our method performs effectively for individual patients, while the cross-patient results suggest its potential for generalization and scalability, helping address real data scarcity challenges through the use of synthetic datasets.
\begin{table}[]
    \centering
    \resizebox{0.9\textwidth}{!}{%
    \begin{tabular}{c|c| *{8}{>{\centering\arraybackslash}p{\rowSize{}}|} *{1}{>{\centering\arraybackslash}p{\rowSize{}}} }
        \toprule
        Metric & Experiments & C1 & C2 & C3 & C4 & C5 & C6 & C7 & C8 & C9 \\
        \midrule
        \rowcolor{\rowColorOne}
        \rowcolor{\rowColorTwo}
        ADD & Patient-specific & 149.0 & 32.75 & 41.27 & 27.13 & 58.23 & 185.8 & 69.95 & 48.54 & 58.58 \\
        \rowcolor{\rowColorThree}
        (mm) & Cross-patient & 67.70 & 46.50 & 33.46 & 54.20 & 96.33 & 132.7 & 34.86 & 53.86 & 32.22 \\
        \midrule
        \rowcolor{\rowColorTwo}
        $E_{rot}$ & Patient-specific & 5.53 & 1.32 & 0.78 & 1.38 & 1.16 & 6.77 & 1.33 & 1.97 & 1.50 \\
        \rowcolor{\rowColorThree}
        ($n^{\circ}$) & Cross-patient & 11.49 & 9.85 & 12.40 & 12.58 & 13.30 & 7.33 & 11.93 & 6.76 & 6.22 \\
        \midrule
        \rowcolor{\rowColorTwo}
        $E_{t}$ & Patient-specific & 150.3 & 33.37 & 41.30 & 27.81 & 59.16 & 184.6 & 69.13 & 46.00 & 58.69 \\
        \rowcolor{\rowColorThree}
        (mm) & Cross-patient & 55.13 & 12.39 & 14.95 & 35.54 & 87.75 & 145.9 & 39.37 & 62.80 & 23.05 \\
        \bottomrule
    \end{tabular}%
    }
    \caption{\textbf{Quantitative Evaluation}. This table presents both patient-specific and cross-patient assessments. C1 represents Case 1.}
    \label{tab:quantitative_evaluation}
\end{table}
Figure~\ref{fig:performance_overall_comparison} shows that patient-specific models achieve higher accuracy in angular distance metric while cross-patient models have more variability. Overall, the majority of cases in both models demonstrate an angular distance below the critical 10-degree threshold \cite{labadie2018preliminary}, suggesting the method’s potential for guiding the electrode array insertion process. While further improvements are needed, these findings lay the groundwork for future clinical applications, where this approach could assist in optimizing insertion angles and reducing the risk of basilar membrane damage.
The qualitative results are presented in Figure~\ref{fig:qualitative_results}, showcasing the overall performance of our proposed method. 
The first row \textit{Original} shows the surgical scenes, while the second row \textit{Pose $\mathbf{P}'$} presents the predicted postmastoidectomy CT mesh poses. The third row \textit{Offset} highlights pose differences using bounding boxes around the ossicles, with ground truth poses in green and predictions in cyan. The final row \textit{Overlay} co-registers key landmarks, including the ossicles (yellow), the facial nerve (magenta), and the chorda (cyan), using $\mathbf{P}'$.
These results demonstrate our model's ability to align 3D anatomical structures with intraoperative 2D imaging, even in complex, cluttered, and occluded surgical scenarios, by directly registering the postmastoidectomy surface to the image. Our proposed 2D-to-3D registration model achieves real-time performance, processing each patient sample at approximately 40 frames per second on an NVIDIA GeForce RTX 4090 GPU. This high frame rate ensures low-latency inference ($\sim$25ms per frame), making it well-suited for continuous intraoperative tracking and efficient surgical navigation.
\newcommand{\FontSize}{\tiny}

\begin{figure}[ht]
    \begin{minipage}[t]{0.45\textwidth}
        \begin{minipage}[c]{0.05\textwidth}
            \centering
            \rotatebox{90}{{\FontSize Original}}
        \end{minipage}
        \begin{minipage}{0.3\textwidth}
            \includegraphics[width=\textwidth]{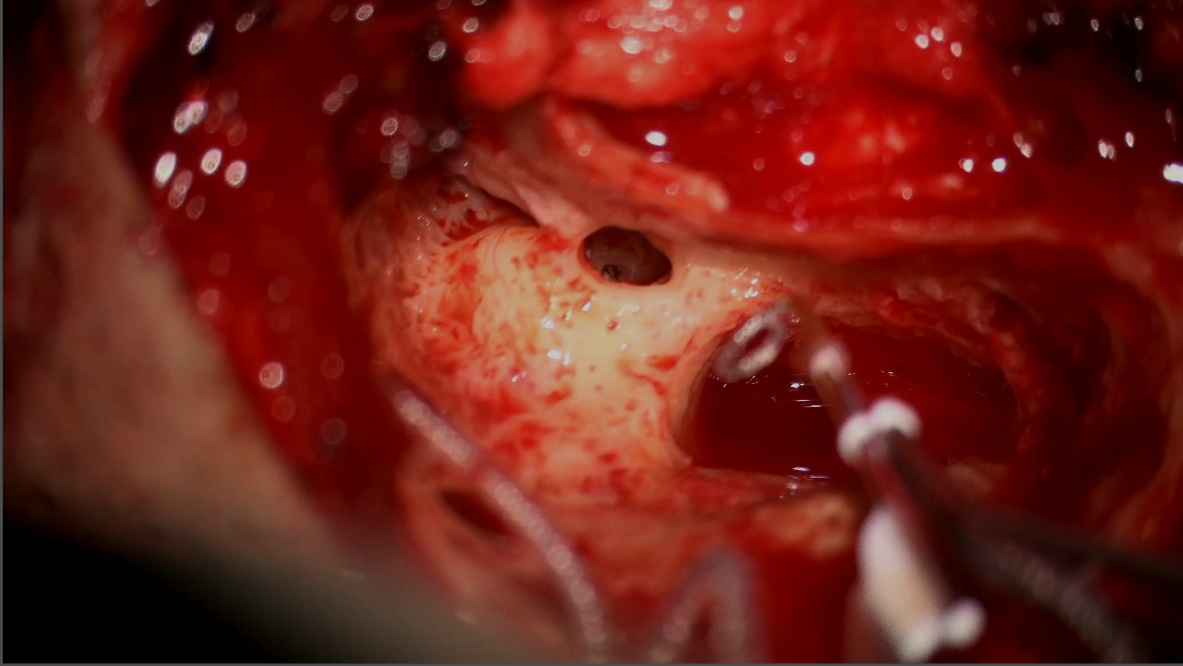}
        \end{minipage}
        \begin{minipage}{0.3\textwidth}
            \includegraphics[width=\textwidth]{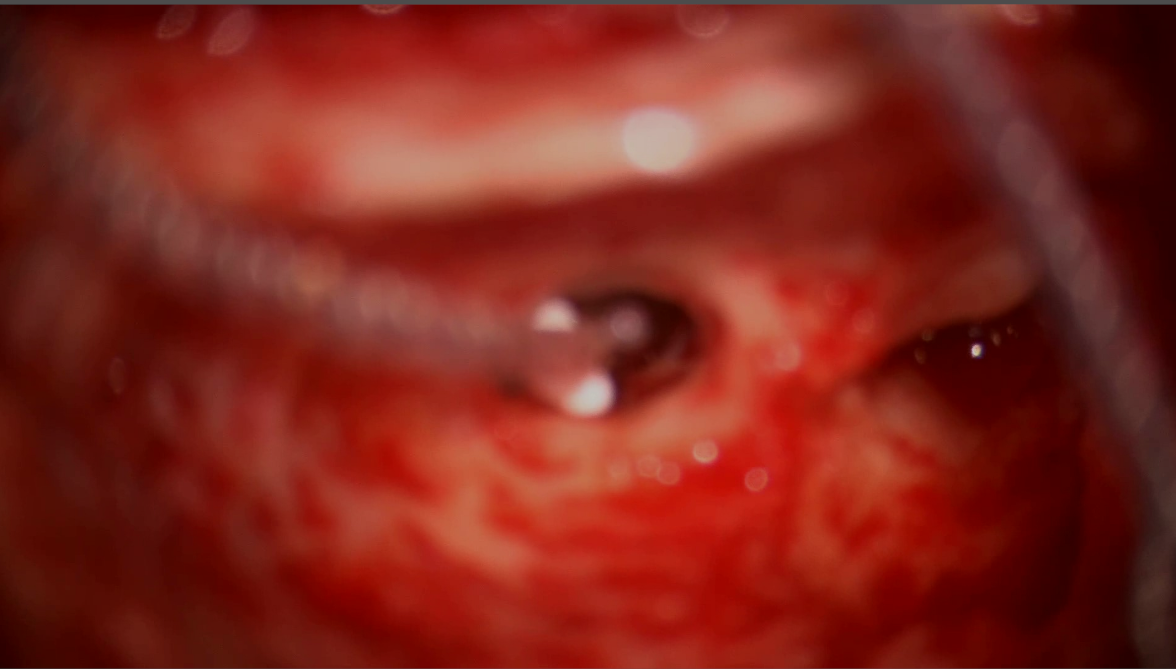}
        \end{minipage}
        \begin{minipage}{0.3\textwidth}
            \includegraphics[width=\textwidth]{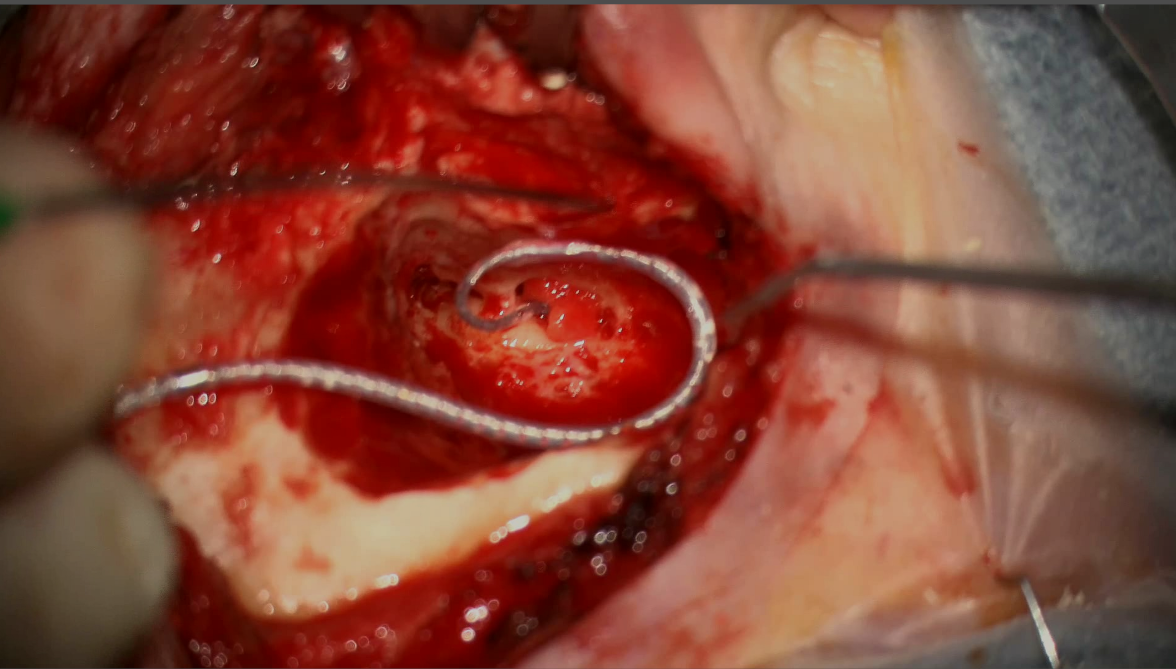}
        \end{minipage}
    \end{minipage}
    \begin{minipage}[t]{0.45\textwidth}
        \begin{minipage}[c]{0.05\textwidth}
            \centering
            \rotatebox{90}{{\FontSize Original}}
        \end{minipage}
         \begin{minipage}{0.3\textwidth}
            \includegraphics[width=\textwidth]{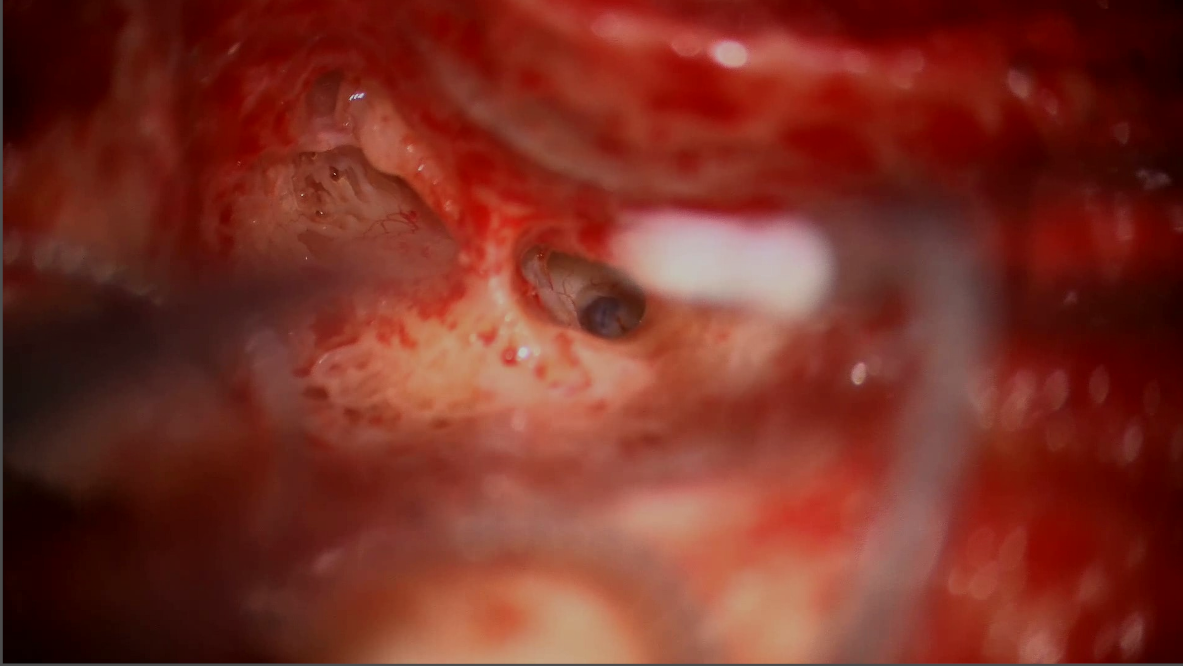}
        \end{minipage}
        \begin{minipage}{0.3\textwidth}
            \includegraphics[width=\textwidth]{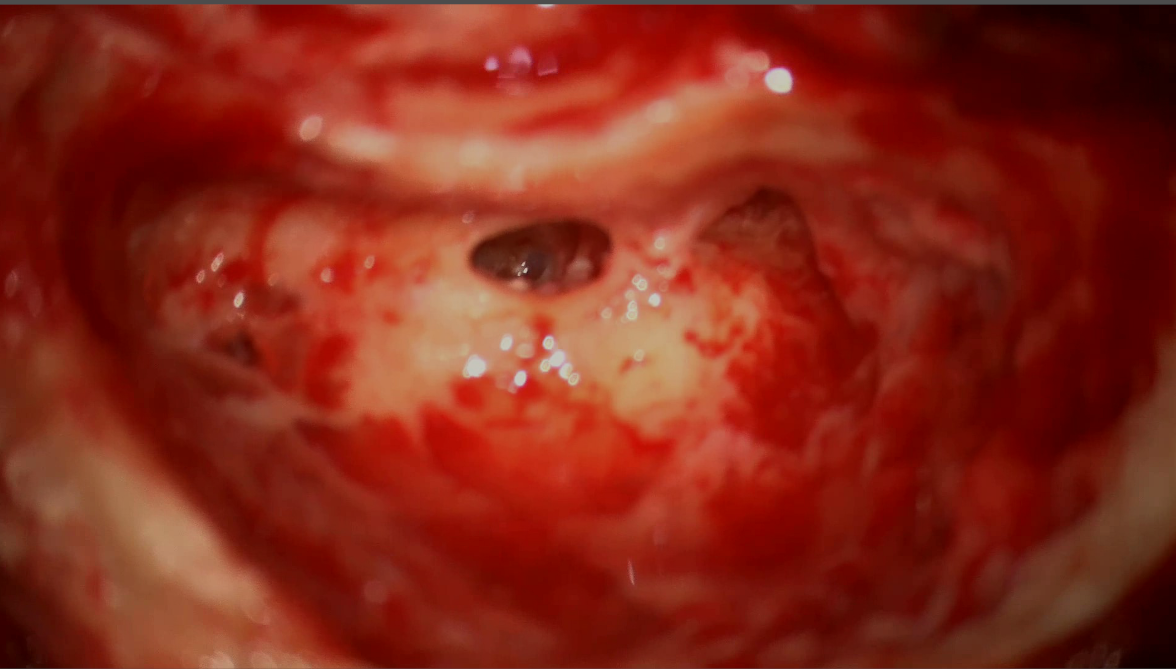}
        \end{minipage}
        \begin{minipage}{0.3\textwidth}
            \includegraphics[width=\textwidth]{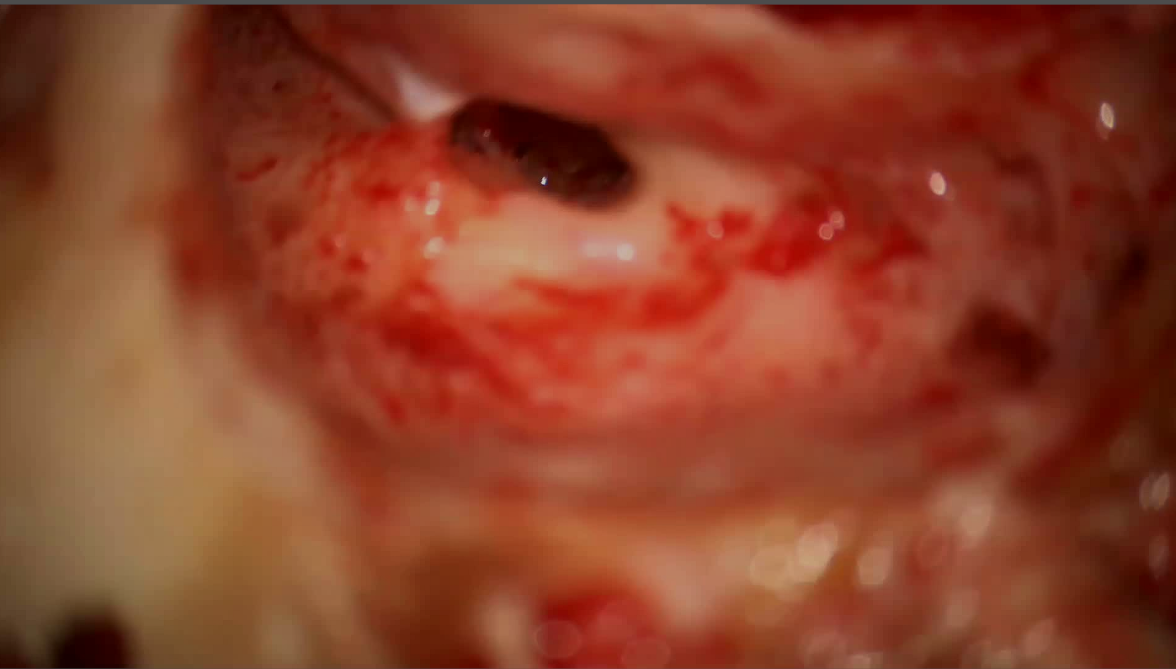}
        \end{minipage}
    \end{minipage}

    \begin{minipage}[t]{0.45\textwidth}
        \begin{minipage}[c]{0.05\textwidth}
            \centering
            \rotatebox{90}{\FontSize Pose $\mathbf{P}'$}
        \end{minipage}
        \begin{minipage}{0.3\textwidth}
            \includegraphics[width=\textwidth]{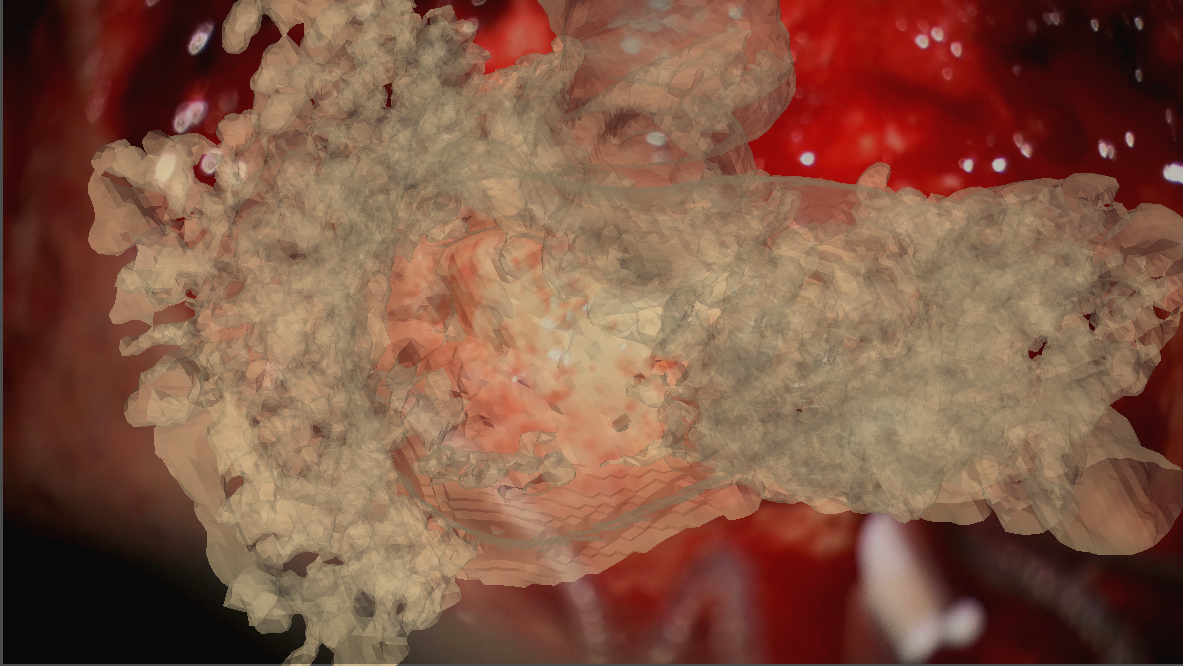}
        \end{minipage}
        \begin{minipage}{0.3\textwidth}
            \includegraphics[width=\textwidth]{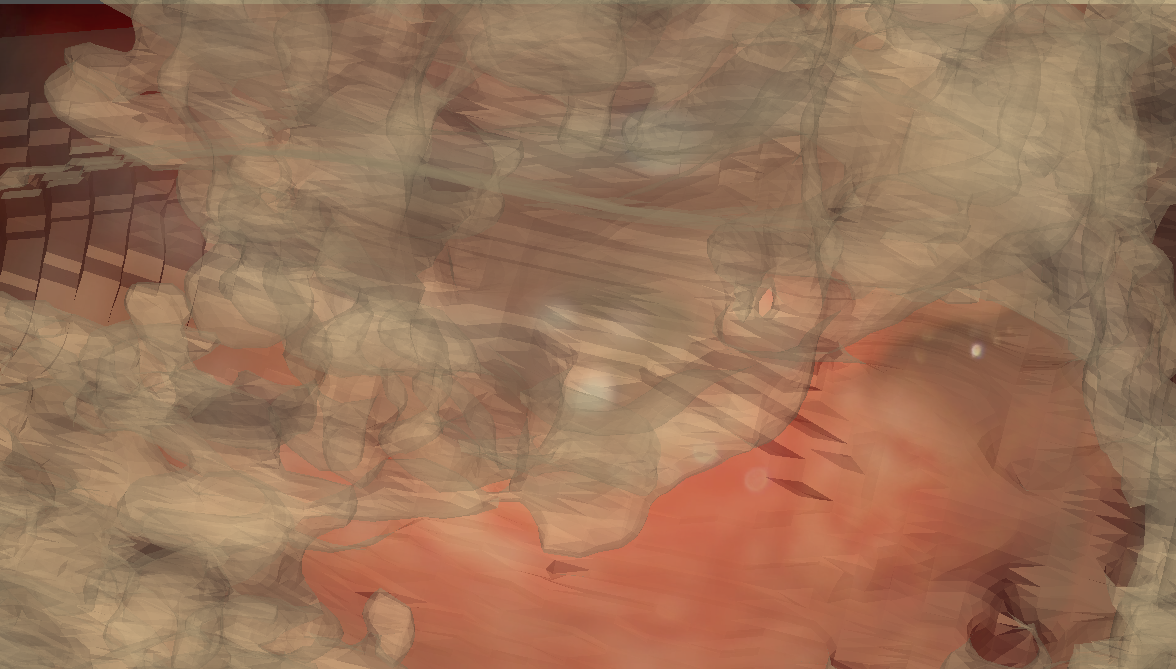}
        \end{minipage}
        \begin{minipage}{0.3\textwidth}
            \includegraphics[width=\textwidth]{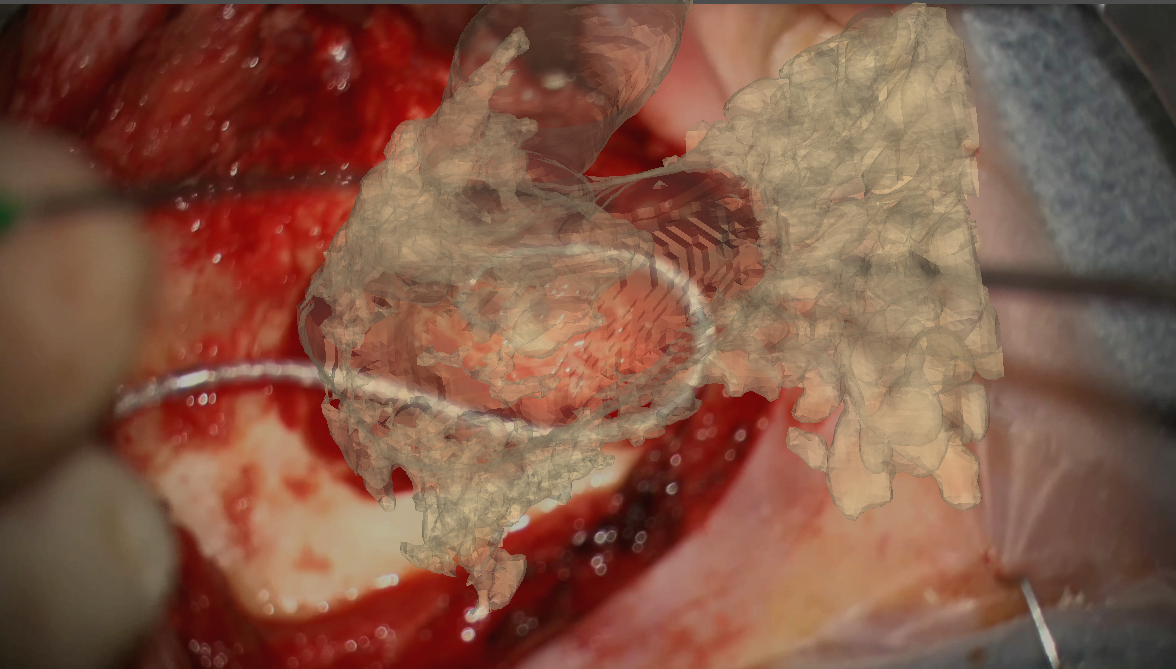}
        \end{minipage}
    \end{minipage}
    \begin{minipage}[t]{0.45\textwidth}
        \begin{minipage}[c]{0.05\textwidth}
            \centering
            \rotatebox{90}{\FontSize Pose $\mathbf{P}'$}
        \end{minipage}
        \begin{minipage}{0.3\textwidth}
            \includegraphics[width=\textwidth]{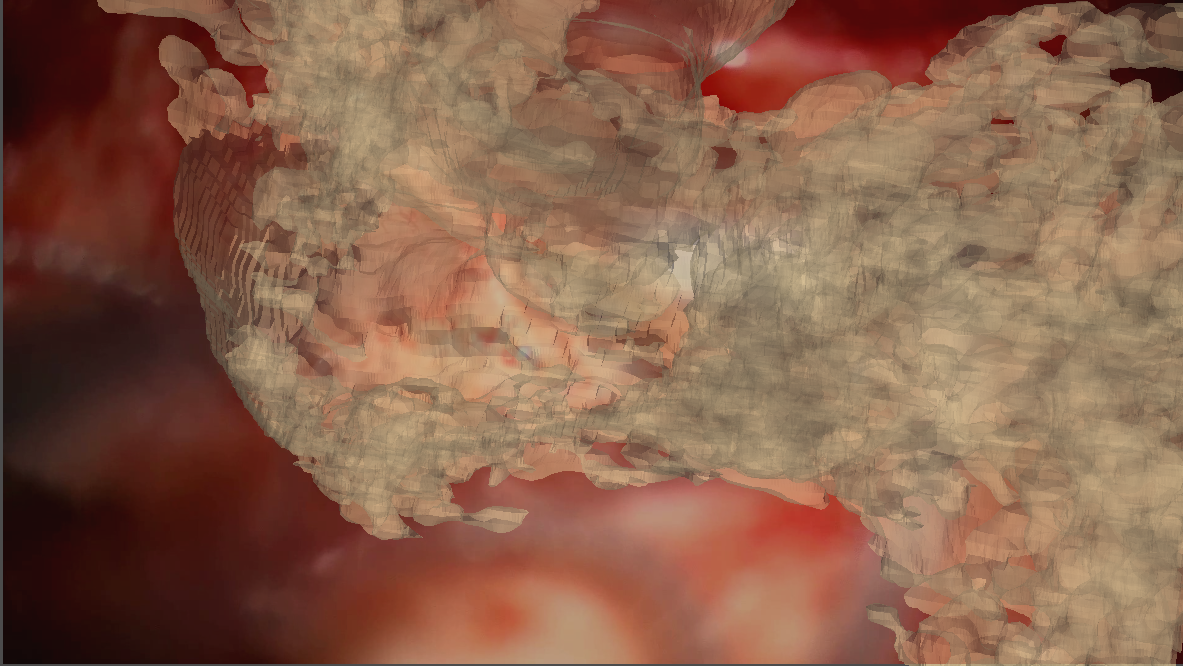}
        \end{minipage}
        \begin{minipage}{0.3\textwidth}
            \includegraphics[width=\textwidth]{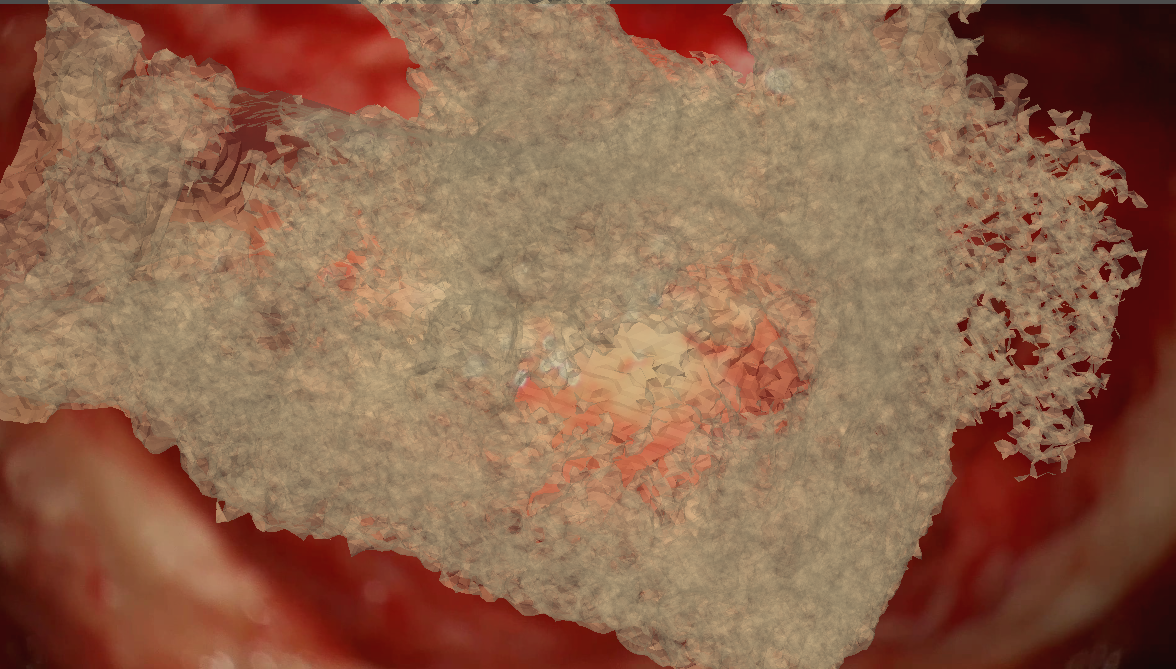}
        \end{minipage}
        \begin{minipage}{0.3\textwidth}
            \includegraphics[width=\textwidth]{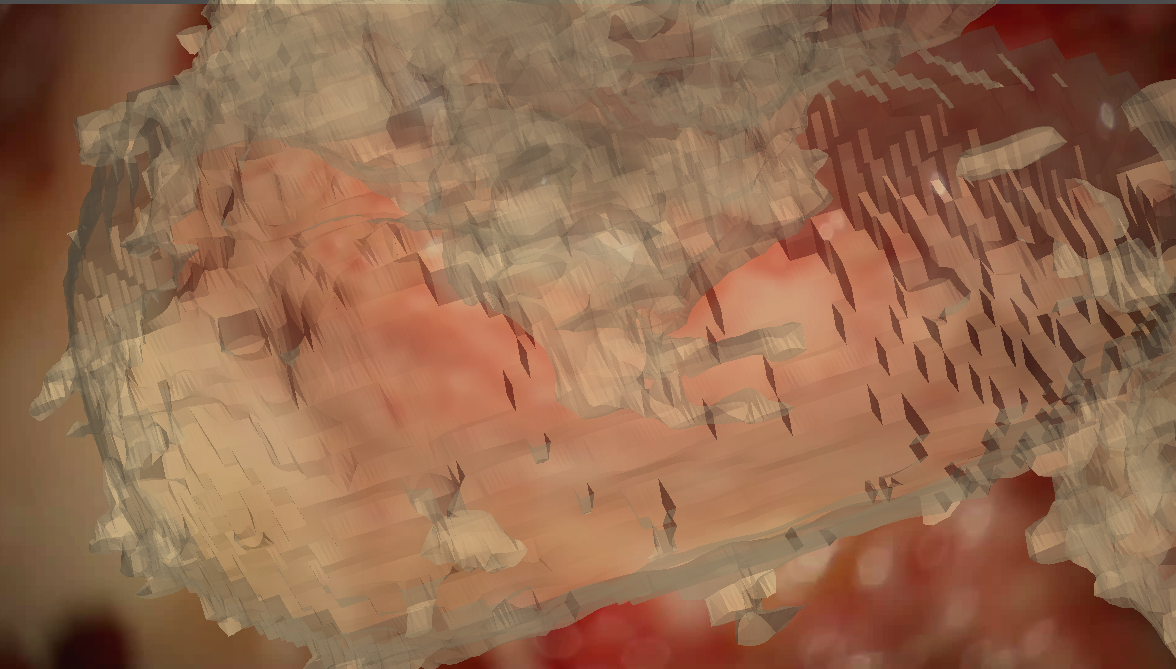}
        \end{minipage}
    \end{minipage}

    \begin{minipage}[t]{0.45\textwidth}
        \begin{minipage}[c]{0.05\textwidth}
            \centering
            \rotatebox{90}{\FontSize Offset}
        \end{minipage}
        \begin{minipage}{0.3\textwidth}
            \includegraphics[width=\textwidth]{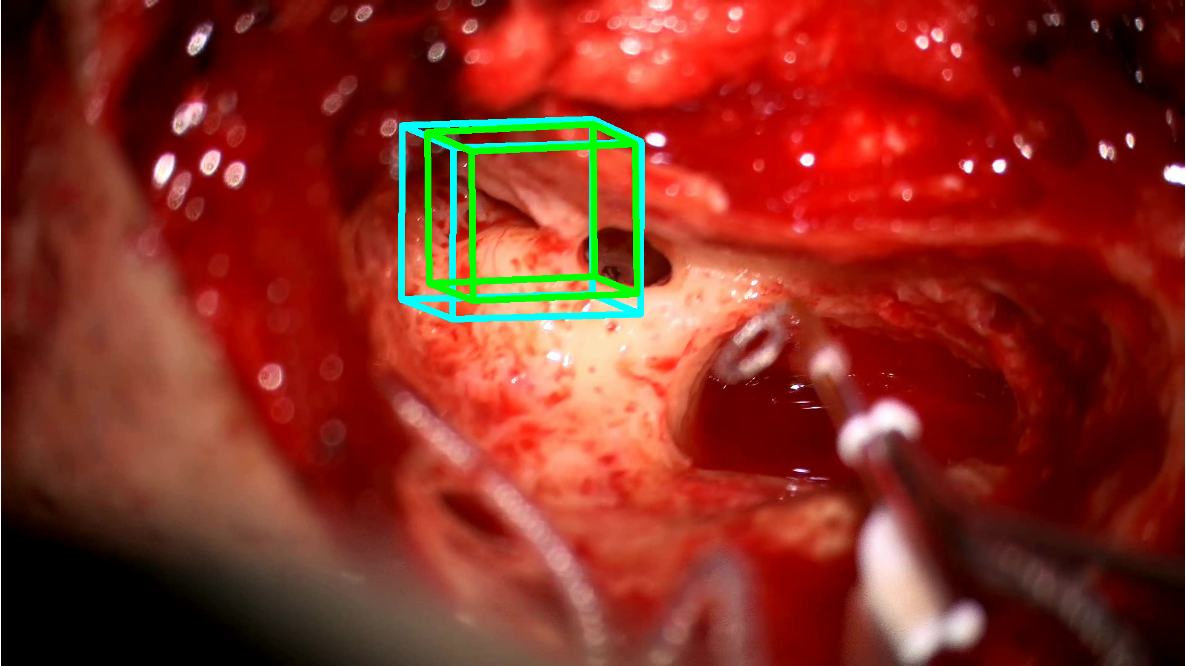}
        \end{minipage}
        \begin{minipage}{0.3\textwidth}
            \includegraphics[width=\textwidth]{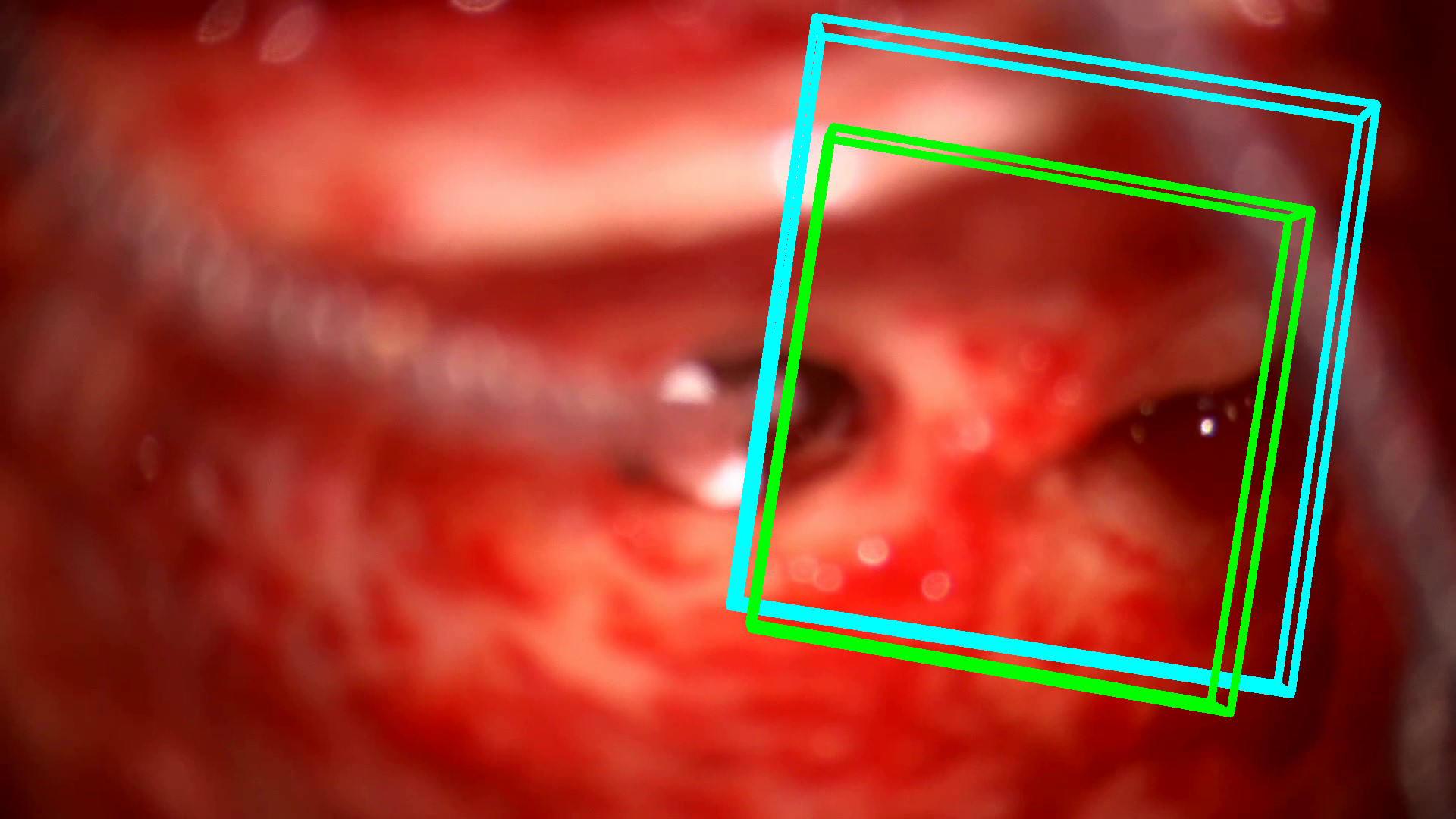}
        \end{minipage}
        \begin{minipage}{0.3\textwidth}
            \includegraphics[width=\textwidth]{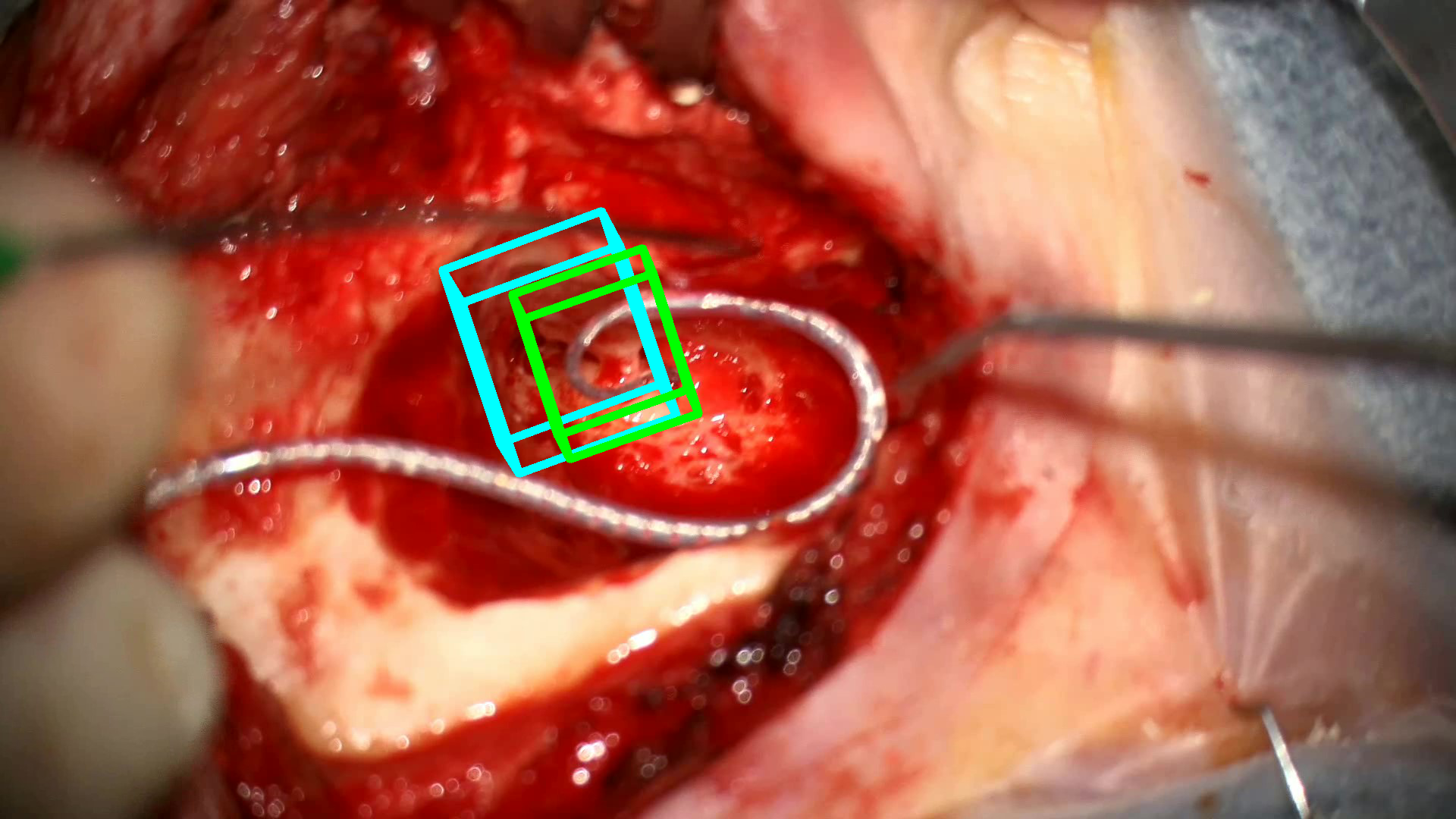}
        \end{minipage}
    \end{minipage}
    \begin{minipage}[t]{0.45\textwidth}
        \begin{minipage}[c]{0.05\textwidth}
            \centering
            \rotatebox{90}{\FontSize Offset}
        \end{minipage}
        \begin{minipage}{0.3\textwidth}
            \includegraphics[width=\textwidth]{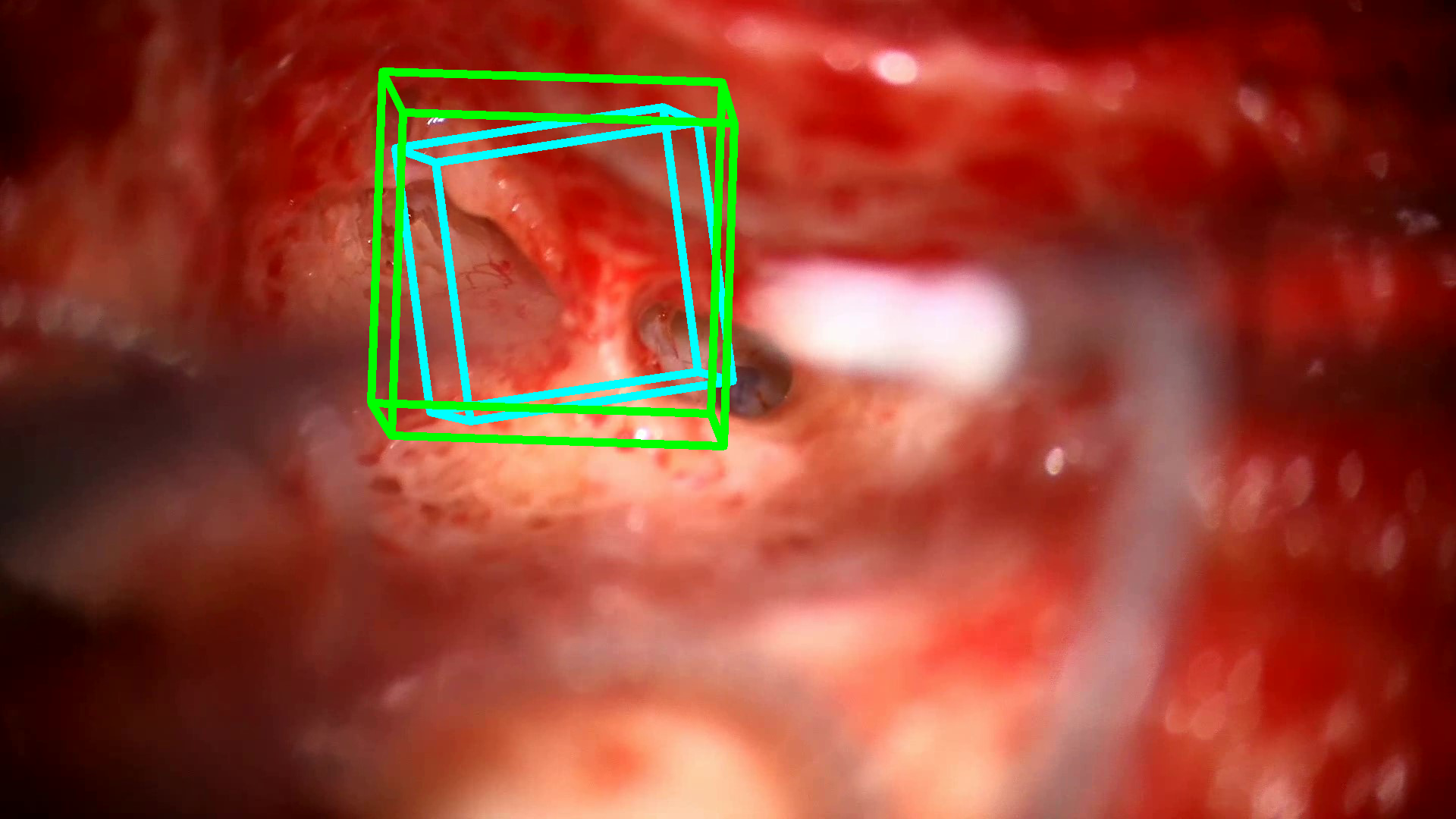}
        \end{minipage}
        \begin{minipage}{0.3\textwidth}
            \includegraphics[width=\textwidth]{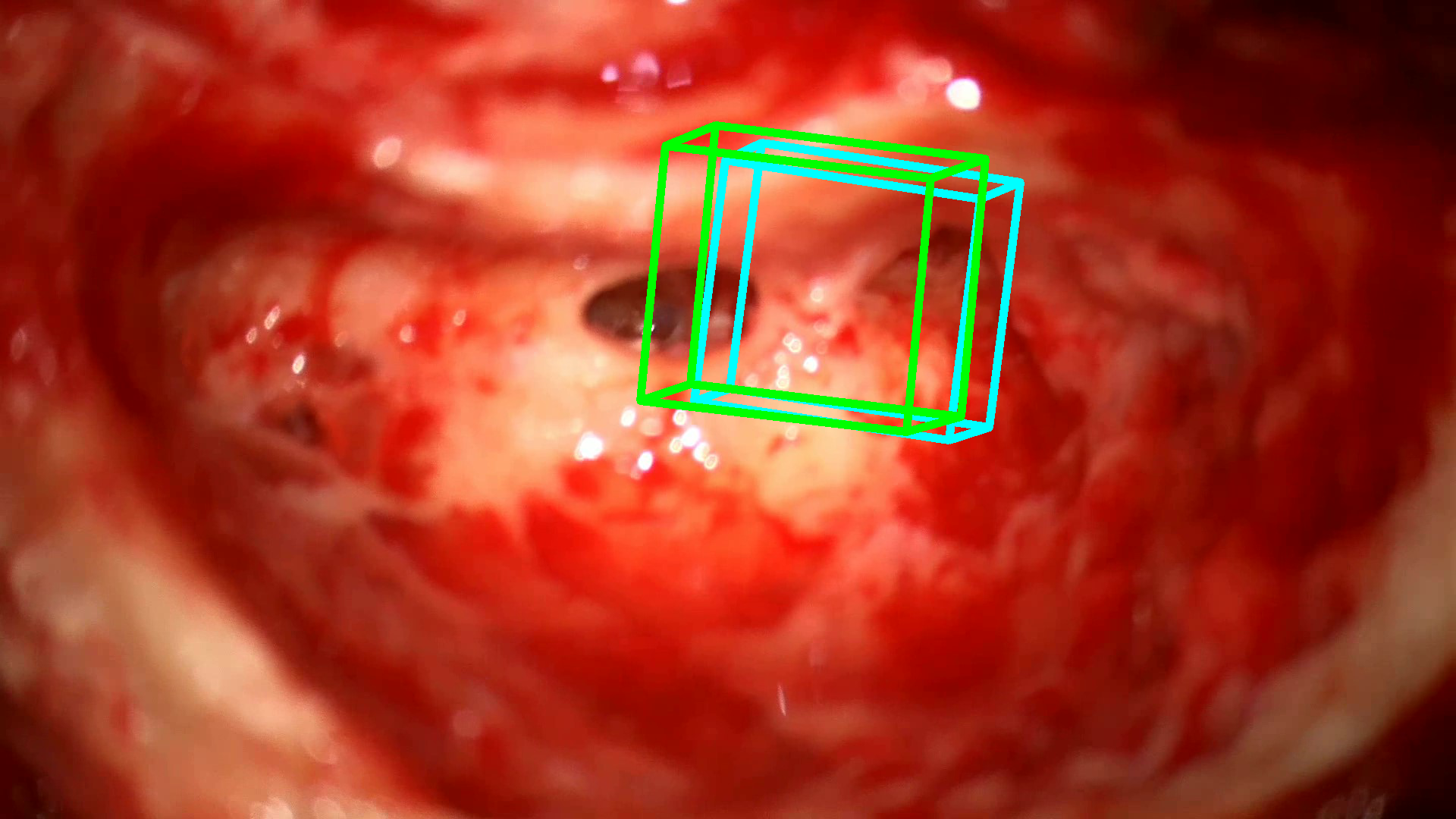}
        \end{minipage}
        \begin{minipage}{0.3\textwidth}
            \includegraphics[width=\textwidth]{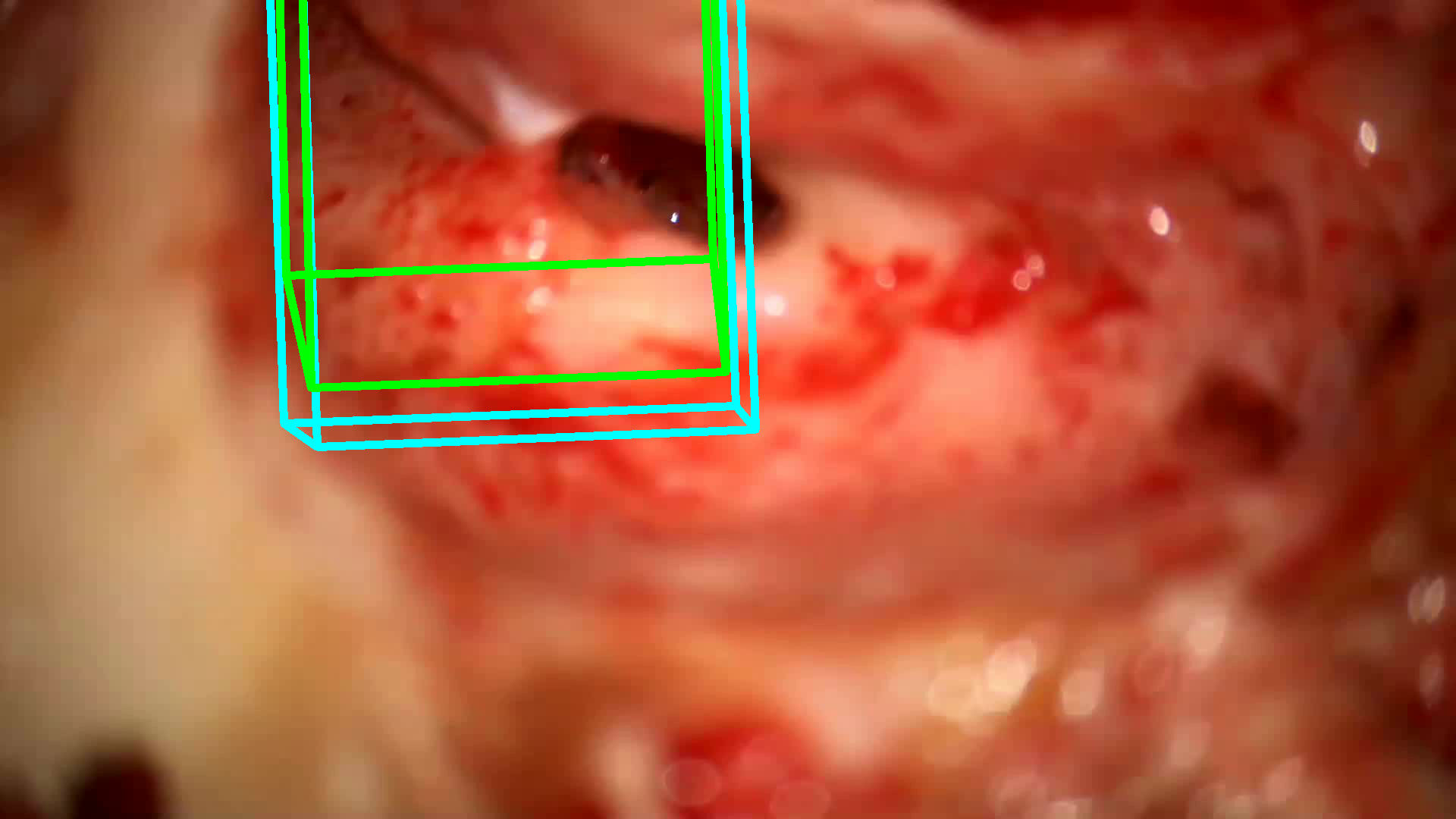}
        \end{minipage}
    \end{minipage}
    
    \begin{minipage}[t]{0.45\textwidth}
        \begin{minipage}[c]{0.05\textwidth}
            \centering
            \rotatebox{90}{\FontSize \phantom{///}Overlay}
        \end{minipage}
        \begin{minipage}{0.3\textwidth}
            \includegraphics[width=\textwidth]{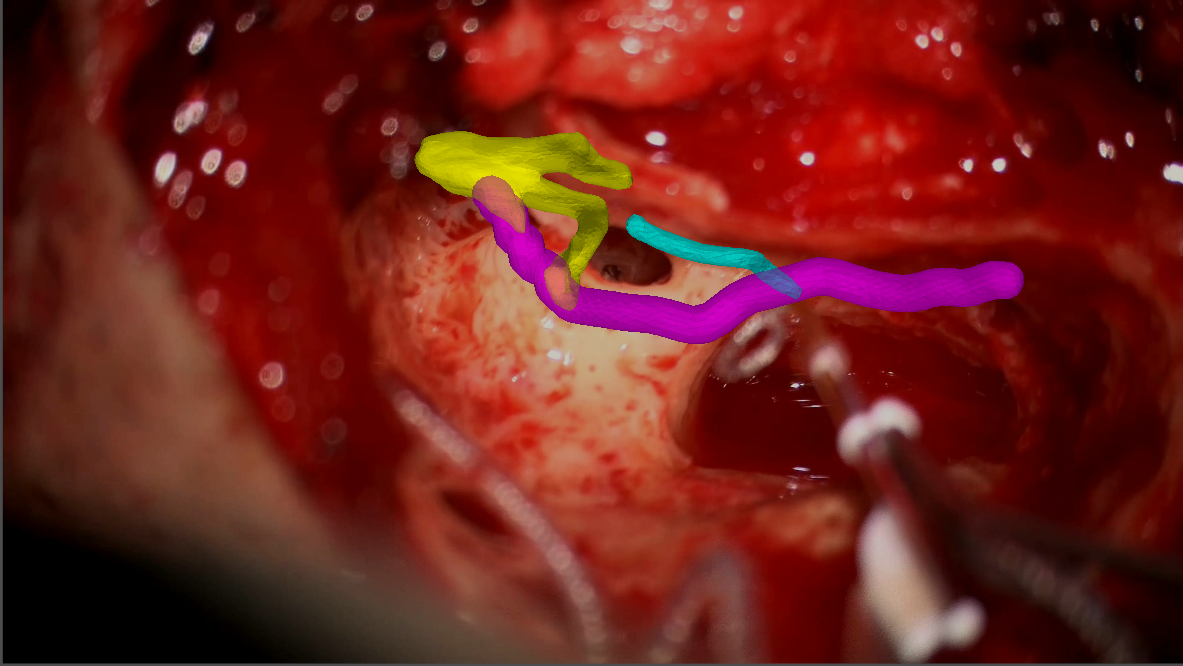}
            \centering
            \footnotesize{Sample 1}
        \end{minipage}
        \begin{minipage}{0.3\textwidth}
            \includegraphics[width=\textwidth]{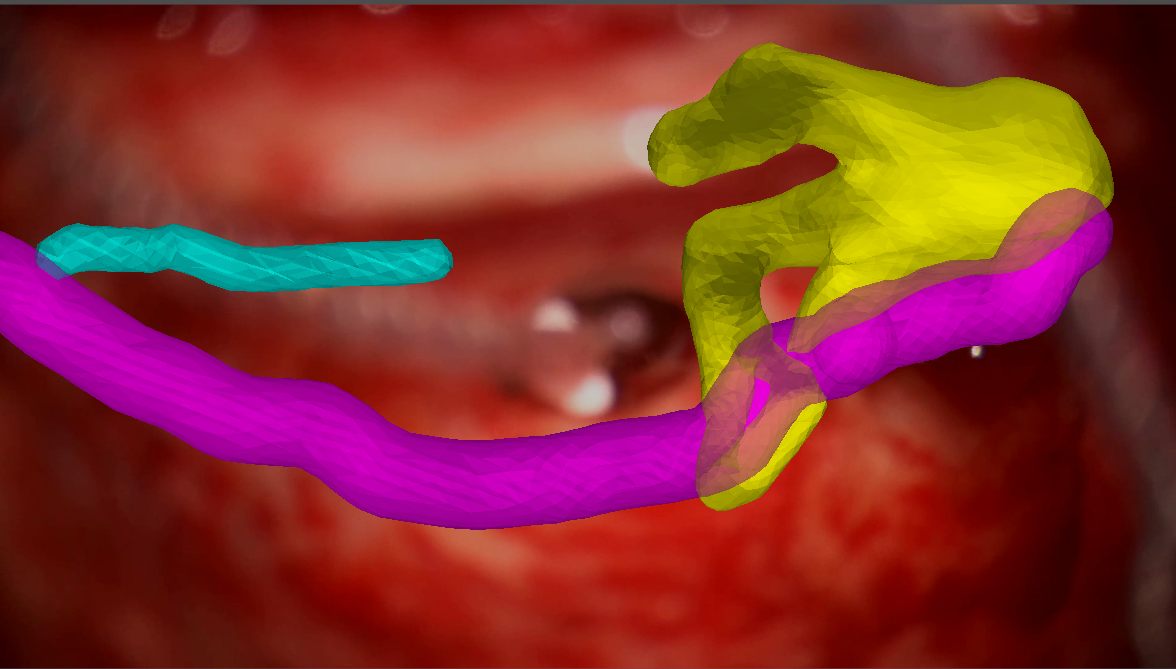}
            \centering
            \footnotesize{Sample 2}
        \end{minipage}
        \begin{minipage}{0.3\textwidth}
            \includegraphics[width=\textwidth]{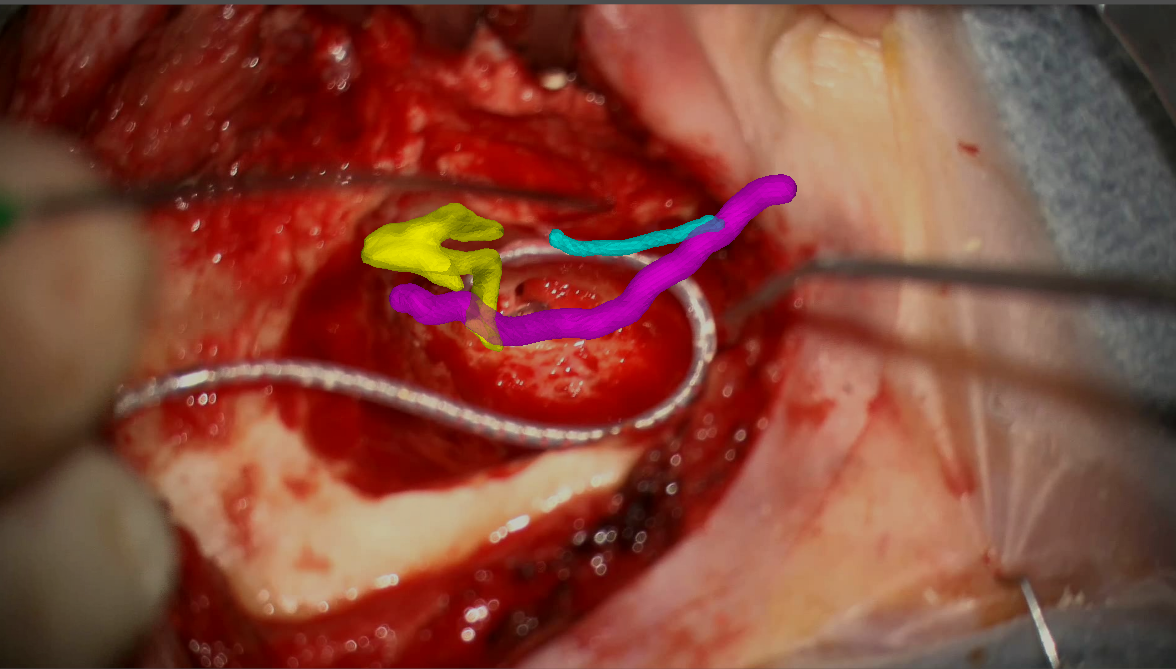}
            \centering
            \footnotesize{Sample 3}
        \end{minipage}
        
        \centering
        Patient-Specific
    \end{minipage}
    \begin{minipage}[t]{0.45\textwidth}
        \begin{minipage}[c]{0.05\textwidth}
            \centering
            \rotatebox{90}{\FontSize \phantom{///} Overlay}
        \end{minipage}
        \begin{minipage}{0.3\textwidth}
            \includegraphics[width=\textwidth]{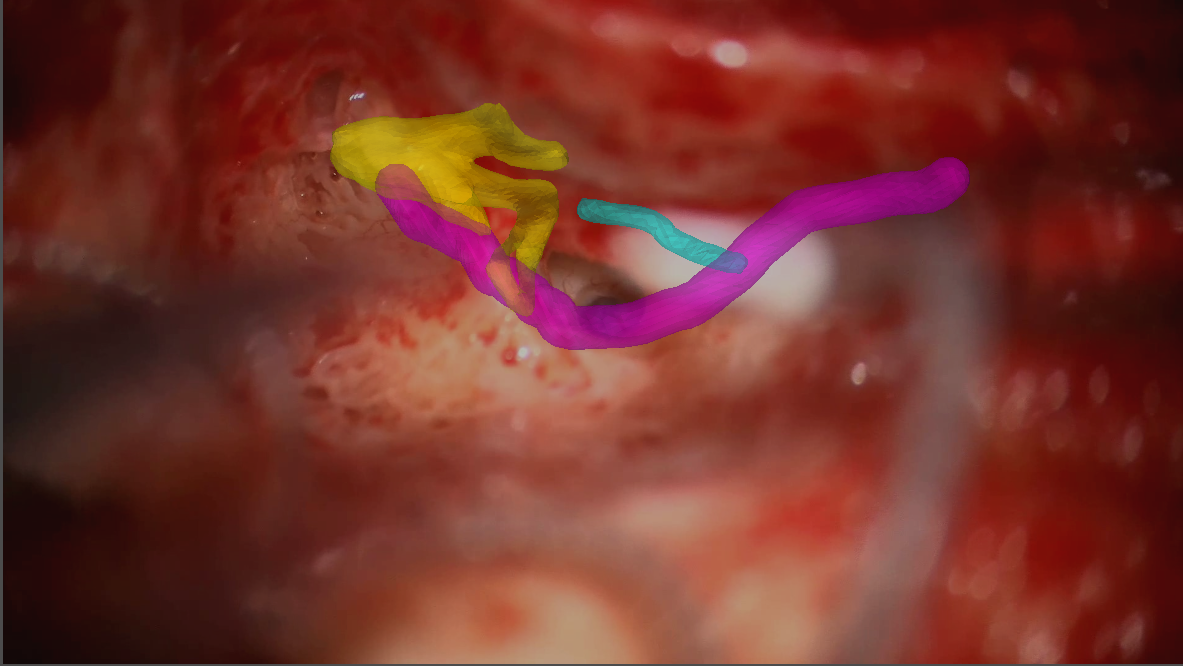}
            \centering
            \footnotesize{Sample 1}
        \end{minipage}
        \begin{minipage}{0.3\textwidth}
            \includegraphics[width=\textwidth]{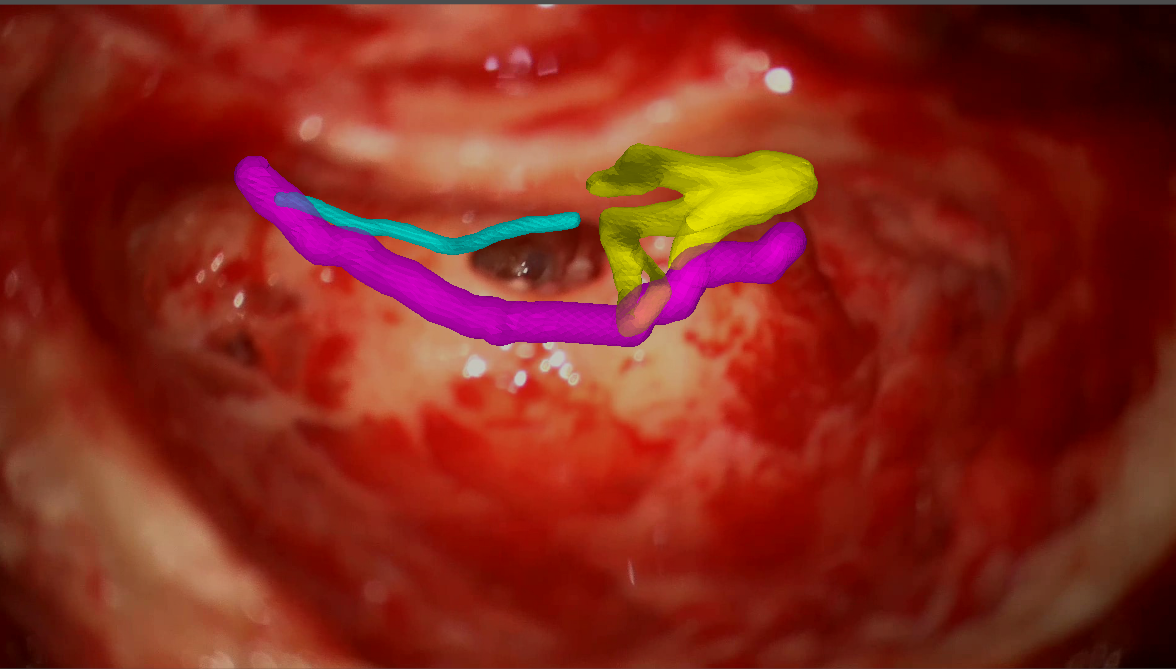}
            \centering
            \footnotesize{Sample 2}
        \end{minipage}
        \begin{minipage}{0.3\textwidth}
            \includegraphics[width=\textwidth]{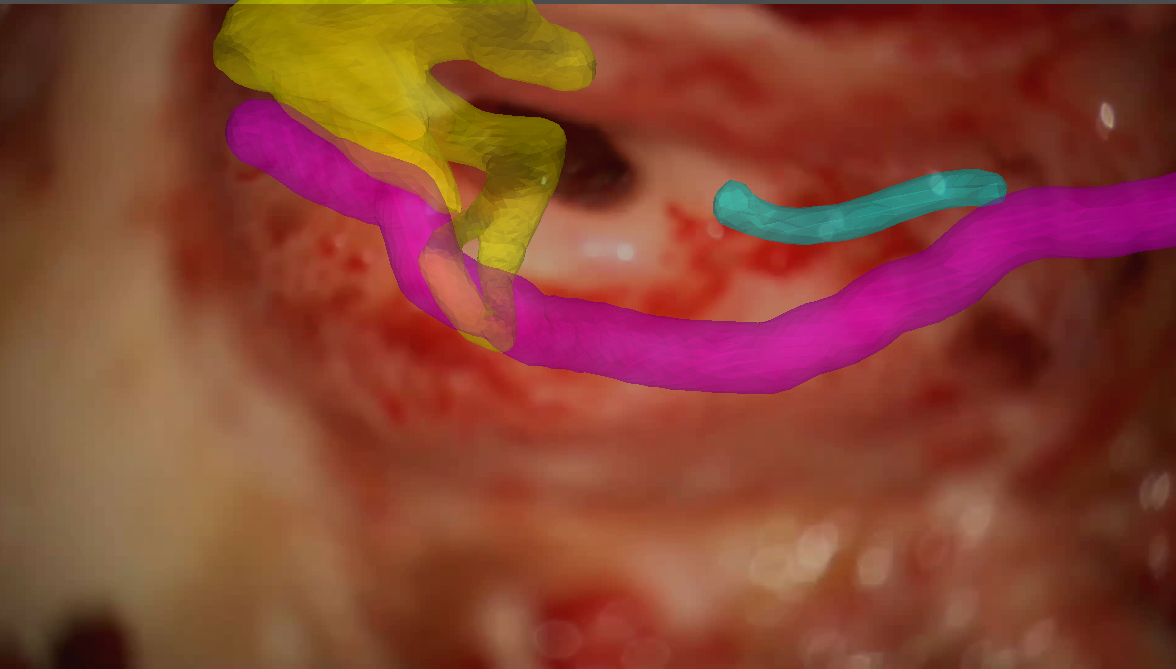}
            \centering
            \footnotesize{Sample 3}
        \end{minipage}
       
        \centering
        Cross-Patient
    \end{minipage}
    \caption{\textbf{Qualitative Evaluation}. We present both the patient-specific and cross-patient results on randomly selected surgical scenes from six different patients.}
  \label{fig:qualitative_results}
\end{figure}
\section{Conclusion}
This paper lays the foundation for marker-free patient-to-image intraoperative registration system in CI surgery, with the potential to optimize electrode array insertion angles and enhance surgical outcomes. However, the proposed method is currently limited to rigid objects in CT scans and does not account for potential deformations in non-rigid anatomical structures. 
While the angular distance error is the primary focus in CI intraoperative navigation for electrode array insertion, integrating a depth estimation model could further reduce z-translation error, improving spatial accuracy in 3D-to-2D alignment. Future work could address these limitations by increasing the training dataset with more patient samples, exploring advanced registration methods, and integrating the proposed model into surgical microscopes for real-time intraoperative guidance.


\bibliography{references.bib}

\begin{thebibliography}{10}

\bibitem{labadie2018preliminary}
Labadie, R. and Noble, J., ``Preliminary results with image-guided cochlear implant insertion techniques,'' {\em Otol Neurotol}~{\bf 39},  922--928 (Aug 2018).

\bibitem{zhang2024mmunsupervisedmambabasedmastoidectomy}
Zhang, Y., Davalos, E., Su, D., Lou, A., and Noble, J.~H., ``M\&m: Unsupervised mamba-based mastoidectomy for cochlear implant surgery with noisy data,'' (2024).

\bibitem{zhang2024mastoidectomymultiviewsynthesissingle}
Zhang, Y. and Noble, J., ``Mastoidectomy multi-view synthesis from a single microscopy image,'' (2024).

\bibitem{zhang2025ssddgansinglestepdenoisingdiffusion}
Zhang, Y., Davalos, E., and Noble, J., ``Ssdd-gan: Single-step denoising diffusion gan for cochlear implant surgical scene completion,'' (2025).

\bibitem{10.1117/12.3008830}
Zhang, Y., Davalos, E., Su, D., Lou, A., and Noble, J.~H., ``{Monocular microscope to CT registration using pose estimation of the incus for augmented reality cochlear implant surgery},'' in [{\em Medical Imaging 2024: Image-Guided Procedures, Robotic Interventions, and Modeling}{\nolinebreak\hspace{0.1em}]},  Siewerdsen, J.~H. and Rettmann, M.~E., eds.,  {\bf 12928},  129282I, International Society for Optics and Photonics, SPIE (2024).

\bibitem{tawfik2024cochlearimplantationslimprecurved}
Tawfik, K.~O., Khan, M. M.~R., Patro, A., Smetak, M.~R., Haynes, D., Labadie, R.~F., Gifford, R.~H., and Noble, J.~H., ``Cochlear implantation of slim pre-curved arrays using automatic pre-operative insertion plans,'' (2024).

\bibitem{lin2023modern}
Lin, Z., Lei, C., and Yang, L., ``Modern image-guided surgery: A narrative review of medical image processing and visualization,'' {\em Sensors (Basel, Switzerland)}~{\bf 23}(24),  9872 (2023).

\bibitem{Furuse2023-un}
Furuse, M., Ikeda, N., Kawabata, S., Park, Y., Takeuchi, K., Fukumura, M., Tsuji, Y., Kimura, S., Kanemitsu, T., Yagi, R., Nonoguchi, N., Kuroiwa, T., and Wanibuchi, M., ``Influence of surgical position and registration methods on clinical accuracy of navigation systems in brain tumor surgery,'' {\em Sci. Rep.}~{\bf 13},  2644 (Feb. 2023).

\bibitem{10.3389/fnbot.2021.636772}
Li, W., Fan, J., Li, S., Tian, Z., Zheng, Z., Ai, D., Song, H., and Yang, J., ``Calibrating 3d scanner in the coordinate system of optical tracker for image-to-patient registration,'' {\em Frontiers in Neurorobotics}~{\bf 15} (2021).

\bibitem{healthcare10101815}
Chiou, S.-Y., Zhang, Z.-Y., Liu, H.-L., Yan, J.-L., Wei, K.-C., and Chen, P.-Y., ``Augmented reality surgical navigation system for external ventricular drain,'' {\em Healthcare}~{\bf 10}(10) (2022).

\bibitem{sahovaler2022automatic}
Sahovaler, A., Daly, M.~J., Chan, H. H.~L., Nayak, P., Tzelnick, S., Arkhangorodsky, M., Qiu, J., Weersink, R., Irish, J.~C., Ferguson, P., and Wunder, J.~S., ``Automatic registration and error color maps to improve accuracy for navigated bone tumor surgery using intraoperative cone-beam ct,'' {\em JB \& JS Open Access}~{\bf 7}(2),  e21.00140 (2022).

\bibitem{taleb2023image}
Taleb, A., Guigou, C., Leclerc, S., Lalande, A., and Bozorg~Grayeli, A., ``Image-to-patient registration in computer-assisted surgery of head and neck: State-of-the-art, perspectives, and challenges,'' {\em Journal of Clinical Medicine}~{\bf 12}(16),  5398 (2023).

\bibitem{haouchine2023learningexpectedappearancesintraoperative}
Haouchine, N., Dorent, R., Juvekar, P., Torio, E., III, W. M.~W., Kapur, T., Golby, A.~J., and Frisken, S., ``Learning expected appearances for intraoperative registration during neurosurgery,'' (2023).

\bibitem{Haouchine2022}
Haouchine, N., Juvekar, P., Nercessian, M., Wells, W., Golby, A., and Frisken, S., ``Pose estimation and non-rigid registration for augmented reality during neurosurgery,'' {\em IEEE Trans. Biomed. Eng.}~{\bf 69},  1310--1317 (Apr. 2022).

\bibitem{zhou2020continuityrotationrepresentationsneural}
Zhou, Y., Barnes, C., Lu, J., Yang, J., and Li, H., ``On the continuity of rotation representations in neural networks,'' (2020).

\bibitem{sullivan2019pyvista}
Sullivan, B. and Kaszynski, A., ``{PyVista}: {3D} plotting and mesh analysis through a streamlined interface for the {Visualization Toolkit} ({VTK}),'' {\em Journal of Open Source Software}~{\bf 4},  1450 (May 2019).

\bibitem{oshea2015introductionconvolutionalneuralnetworks}
O'Shea, K. and Nash, R., ``An introduction to convolutional neural networks,'' (2015).

\bibitem{dropout}
Srivastava, N., Hinton, G., Krizhevsky, A., Sutskever, I., and Salakhutdinov, R., ``Dropout: A simple way to prevent neural networks from overfitting,'' {\em Journal of Machine Learning Research}~{\bf 15}(56),  1929--1958 (2014).

\bibitem{xu2015empiricalevaluationrectifiedactivations}
Xu, B., Wang, N., Chen, T., and Li, M., ``Empirical evaluation of rectified activations in convolutional network,'' (2015).

\bibitem{chen2024gridmaskdataaugmentation}
Chen, P., Liu, S., Zhao, H., Wang, X., and Jia, J., ``Gridmask data augmentation,'' (2024).

\bibitem{6619221}
Shotton, J., Glocker, B., Zach, C., Izadi, S., Criminisi, A., and Fitzgibbon, A., ``Scene coordinate regression forests for camera relocalization in rgb-d images,'' in [{\em 2013 IEEE Conference on Computer Vision and Pattern Recognition}{\nolinebreak\hspace{0.1em}]},   2930--2937 (2013).

\end{thebibliography}
\bibliographystyle{spiebib}
\end{document}